\documentclass[lettersize,journal]{IEEEtran}
\usepackage[ruled,linesnumbered]{algorithm2e}
\usepackage{amsmath,amsfonts}
\usepackage{array}
\usepackage[caption=false,font=normalsize,labelfont=sf,textfont=sf]{subfig}
\usepackage{textcomp}
\usepackage{stfloats}
\usepackage{url}
\usepackage{verbatim}
\usepackage{graphicx}
\usepackage{cite}
\usepackage{float}
\usepackage{tabularx}
\usepackage{booktabs,makecell, multirow, tabularx}
\usepackage{titlesec}
\titlespacing{\subsection}{0pt}{9pt}{3pt}
\titlespacing{\subsubsection}{0pt}{0pt}{0pt}
\usepackage{hyperref}
\usepackage{etoolbox}

\hypersetup{
	colorlinks=true,
	linkcolor=blue,
	filecolor=blue,      
	urlcolor=blue,
	citecolor=blue,
}
\begin{document}

\title{MPCM-Net: A multi-scale network that integrates partial attention convolution with Mamba for ground-based cloud image segmentation}

\author{{Penghui Niu, Jiashuai She, Taotao Cai, Yajuan Zhang, Ping Zhang, Junhua Gu, and Jianxin Li, \textit{Senior Member, IEEE}}

\thanks{This work was supported in part by The National Natural Science Foundation of China under Grant No. 62206085, in part by the Innovation Capacity Improvement Plan Project of Hebei Province under 22567603H, and in pary by the Interdisciplinary postgraduate Training Program of Hebei University of Technology under HEBUT-Y-XKJC-2022101. \textit{(Corresponding author: Junhua Gu.)}}
\thanks{Penghui Niu and Yajuan Zhang are with the School of Artificial Intelligence, Hebei University of Technology, Tianjin 300401, China (e-mail: qingxinqazxsw@163.com; zhangyajuan@scse.hebut.edu.cn).

Jiashuai She is with the School of Electrical Engineering, Hebei University of Technology, Tianjin 300401, China (1004862447@qq.com).

Taotao Cai is with the University of Southern Queensland, Toowoomba 487-
535, Australia (e-mail: taotao.cai@usq.edu.au).

Ping Zhang and Junhua Gu is with the the School of Artificial Intelligence, Hebei University of Technology, Tianjin 300401, China, and also with the Hebei Province Key Laboratory of Big Data Calculation, Hebei University of Technology, Tianjin 300401, China (e-mail: jhguhebut@163.com; zhangping@hebut.edu.cn).

Jianxin Li is with the Discipline of Business Systems and Operations, School of Business and Law, Edith Cowan University, Joondalup, WA 6027, Australia (e-mail: jianxin.li@ecu.edu.au).
}}

\markboth{Journal of \LaTeX\ Class Files,~Vol.~14, No.~8, August~2021}%
{Shell \MakeLowercase{\textit{et al.}}: A Sample Article Using IEEEtran.cls for IEEE Journals}


\maketitle
\begin{abstract}
Ground-based cloud image segmentation is a critical research domain for photovoltaic (PV) power forecasting. Current deep learning (DL) approaches primarily focus on encoder-decoder architectural refinements. However, existing methodologies exhibit several limitations: (1) they rely on dilated convolutions for multi-scale context extraction, yet fail to leverage inter-channel interoperability and partial feature efficacy; (2) implementations of attention-based feature enhancement frequently compromise the equilibrium between accuracy and throughput; and (3) the decoder modifications often fail to re-establish global interdependencies among hierarchical local features, thereby constraining inference efficiency. To mitigate these challenges, we propose MPCM-Net, a \underline{M}ulti-scale network that integrates \underline{P}artial attention \underline{C}onvolutions with \underline{M}amba architectures to enhance segmentation accuracy. Specifically, the encoder incorporates a multi-scale partial attention convolution (MPAC), which comprises: (1) a multi-scale partial convolution block (MPC) with partial channel module (ParCM) and partial spatial module (ParSM) that facilitating global spatial interaction across multi-scale cloud formations, and (2) a multi-scale partial attention block (MPA) combining partial attention module (ParAM) and ParSM to extract discriminative features with reduced computational complexity. On the decoder side, a multi-scale Mamba block (M2B) is employed to mitigate contextual loss through a spatial-semantic hybrid domain (SSHD) that maintains linear complexity while enabling deep feature aggregation across spatial and scale dimensions. Furthermore, we introduce and release a dataset incorporating \textit{Complex-Scale variations, Radiative properties, and Color attributes (CSRC)}, which is a clear-label, fine-grained segmentation benchmark designed to overcome the critical limitations of existing public datasets. Extensive empirical analysis on CSRC demonstrates the superior performance of MPCM-Net over state-of-the-art methods, achieving an optimal balance between segmentation accuracy and inference speed. The dataset and source code will be available at https://github.com/she1110/CSRC.
\end{abstract}


\section{Introduction}\label{sec:intro}
\IEEEPARstart{C}{louds} constitute critical atmospheric components governing solar radiation transmission and energy balance, directly determining surface insolation intensity with profound implications for agricultural irrigation, aviation safety, and renewable energy forecasting \cite{bg}. Contemporary cloud monitoring technologies are broadly categorized into satellite remote sensing \cite{satellite} and ground-based observations \cite{PV}, with the latter offering superior spatiotemporal resolution (updating at second-scale intervals) and minimized atmospheric attenuation at low observation altitudes. This enables the precise capture of cloud morphology, textural features, and kinematic trajectories. Particularly in PV power forecasting, precisely segmented ground-based cloud imagery facilitates real-time identification of cloud formations, providing minute-resolution irradiance fluctuation alerts for solar farms. Consequently, it significantly improves power generation forecast accuracy while delivering unique value for grid dispatch operations and clean energy integration.

Advancing the state-of-the-art (SOTA) of accurate ground-based cloud image segmentation requires overcoming two core technical challenges: a) developing models with the capacity to characterize multi-scale morphological variations within individual cloud formations across sequential frames, and b) implementing computationally efficient algorithms that precisely capture intricate boundary details while satisfying real-time inference constraints for operational deployment.

Existing methods have sought to address these challenges through two primary avenues: approaches based on image processing with handcrafted features, and the more recently emerging DL-based methods. The former category, which includes threshold-based \cite{threshold} and spatial feature-based approaches \cite{spatial}, typically entails high computational costs and requires extensive manual annotation, resulting in suboptimal segmentation of cloud imagery details. Methods that rely on engineered features are often hindered in their ability to capture the multi-scale characteristics of cloud formations, thereby restricting model accuracy and generalization ability \cite{ML}.

In contrast, DL methodologies have gained prominence due to their superior automated feature extraction capabilities. Classical semantic segmentation architectures, such as U-Net \cite{U-Net}, DeepLab\cite{DeepLabv3}, and SegNet\cite{SegNet}, have demonstrated significant efficacy in ground-based cloud segmentation tasks\cite{CloudSegNet, SegCloud}. To specifically address the multi-scale nature of cloud formations, researchers have integrated multi-scale dilated convolutions within encoders, effectively expanding receptive fields without loss of resolution\cite{CloudU-Net, MA-SegCloud}. Concurrently, attention mechanisms have been introduced to dynamically re-weight features; by prioritizing salient information, these mechanisms facilitate the capture of long-range contextual dependencies, thereby refining multi-scale feature representation\cite{Att1, Att2}. Ideally, the decoder should recover spatial details lost during the encoding process. However, standard upsampling operators often introduce artifacts or fail to reconstruct fine-grained boundary details. To mitigate this contextual information loss, significant research has focused on advanced, learnable upsampling operators\cite{CloudU-Netv2, CloudDeepLabV3+}. Simultaneously, auxiliary modules and Self-Attention (SA) mechanisms have been integrated into decoders to aggregate multi-scale features and reinforce global representations\cite{ViT, SViT}.

Despite these encoder and decoder refinements improving segmentation performance, several limitations persist. Firstly, the multi-scale dilated convolution methods employed in encoders often overlook the efficacy and interconnectivity of partial spatial and channel features, impairing global information interaction. Moreover, the implemented attention mechanisms frequently disregard the computational overhead and latency implications for the network model. Furthermore, decoder enhancements commonly neglect the correlation between global features and local information across different hierarchical levels, as well as the critical trade-off between segmentation accuracy and inference efficiency. 

Accurate segmentation of ground-based cloud images represents a critical task in PV power forecasting. However, most publicly available datasets currently only support binary segmentation between clouds and sky background, which hinders the accurate prediction of PV power generation. This limitation primarily originates from the fact that radiation sources and cloud layers of varying colors severely restrict the utility of these segmentation maps for accurate, real-world PV power forecasting, which is highly dependent on cloud optical properties. Consequently, the development of fine-grained cloud image segmentation datasets and techniques that incorporate radiation sources and color categorization is particularly imperative.

To address these distinct challenges, this paper proposes the Multi-scale network that integrates partial attention convolution with Mamba (MPCM-Net), a novel architecture that integrates multi-scale partial attention convolution in the encoder and multi-scale Mamba in the decoder. Specifically, at the encoder stage, we introduce a multi-scale partial attention convolution (MPAC) to adaptively extract multi-scale contextual information while enhancing feature interaction. The MPAC comprises two key components: 1) a multi partial convolution block (MPC) incorporating partial channel module (ParCM) and partial spatial module (ParSM) is designed to facilitate interaction between partial and global information across diverse cloud formations; and 2) a multi partial attention block (MPA) combining partial attention module (ParAM) and ParSM, which employs a novel attention mechanism to expand the network's receptive field and improve multi-scale feature perception. For the decoder part, a multi-scale Mamba module (M2B) is implemented to mitigate boundary detail loss during upsampling. M2B introduces a spatial-semantic hybrid domain (SSHD) that enhances feature representation across spatial and cross-scale dimensions, thereby achieving an effective balance between segmentation accuracy and inference speed. Finally, we develop and release a new dataset that incorporates fine-grained segmentation based on radiative sources and color attributes. The main contributions of this article are summarized as follows.
\begin{enumerate}
	\item{A novel multi-scale network for ground-based cloud image segmentation, MPCM-Net, is proposed. It explicitly integrates multi-scale partial attention convolution with Mamba architecture to enhance multi-scale feature extraction, thereby strengthening boundary feature representation while significantly improving segmentation accuracy.}
	\item{A multi-scale partial attention convolutional is introduced in the encoder. It features a multi partial convolution block that leverages partial channel and spatial information to facilitate global information exchange across diverse cloud formations and a multi partial attention block to optimize multi-scale feature perception while maintaining computational efficiency.}
	\item{A multi Mamba block is designed in the decoder. It introduces an SSHD to capture hierarchical features, enhancing the capability of obtaining global correlations within local multi-scale information and effectively mitigating contextual information loss.}
	\item{The introduction and public release of the CSRC dataset. As a major contribution to the community, the CSRC dataset is a newly introduced complex-scale, clear-label, and fine-grained segmentation benchmark that addresses critical limitations of prior datasets. It establishes a more realistic and challenging foundation for advancing ground-based cloud image segmentation research. Extensive experiments on CSRC demonstrate that MPCM-Net outperforms SOTA methods in both prediction accuracy and computational efficiency.}
\end{enumerate}

The rest of this paper is organized as follows. In Section~\ref{sec:related}, we introduce several studies pertaining to improving the encoder and decoder in ground-based cloud image segmentation. In Section~\ref{sec:method}, we introduce the detailed structure of the proposed method. In Section~\ref{sec:exp}, we evaluate the performance of our proposed methods. Finally, Section~\ref{sec:conclusion} concludes the paper.

\section{Related Works} \label{sec:related}
Accurate segmentation of ground-based cloud imagery holds critical significance for climate change studies, environmental monitoring, and weather forecasting, garnering significant scholarly interest in recent years. Methodologies in this domain are generally categorized into traditional algorithms and DL-based methods. Traditional algorithms, encompassing threshold-based, spatial feature-based, and machine learning-based approaches, are frequently constrained by limited accuracy and generalizability due to their dependency on hand-crafted feature engineering. Conversely, with rapid advances in computer vision, DL-based methods have achieved remarkable performance by employing encoder-decoder architectures that automatically learn hierarchical representations.
\subsection{Encoder for Ground-based Cloud Image Segmentation}
The encoder fundamentally serves to extract features and compress information, transforming raw inputs into semantically rich representations. While foundational architectures like U-Net\cite{U-Net} and DeepLabv3+\cite{DeepLabv3} have been widely adopted, recent research focuses on enhancing the encoder's capacity to handle the extreme scale variations and complex boundaries of cloud formations \cite{Gonzales, Lu}. Strengthening the encoder's ability to capture multi-scale features is paramount. Early approaches integrated dilated convolutions into U-Net-like structures to expand receptive fields without losing spatial resolution \cite{CloudU-Net, DDUNet}. To accommodate the irregular geometric morphology of clouds, Zhang et al. introduced multi-branch asymmetric convolution modules \cite{MA-SegCloud}. These mechanisms utilize non-square convolution kernels to better capture directional features and employ adaptive channel weighting to refine contextual information. Similarly, Li et al. have proposed heterogeneous receptive field enhancement strategies, which combine depthwise separable convolutions with varied dilation rates to extract dense multi-scale features efficiently \cite{CloudDeepLabV3+}. More recently, dynamic multi-scale convolution mechanisms constitute a significant advancement; in contrast to static filters, these modules adaptively expand the receptive field based on the input image properties, significantly improving the network's perception of variable cloud structures \cite{CloudFU-Net}. Standard convolutional operations often struggle to model long-range dependencies in large-scale cloud systems. To mitigate this, strip pooling operations have been adopted to capture global context along narrow axes, which is particularly effective for segmenting streak-like cloud patterns \cite{MFAFNet}. Additionally, Hu et al. explored the incorporation of embedding layers to enhance regional recognition capabilities \cite{FA-CloudSeg}. With the prevalence of attention mechanisms, hybrid architectures integrating CNNs and Transformers have emerged. Shi et al. have designed fine-grained feature fusion modules to integrate local details with global context \cite{CloudSwinNet}. Hybrid CNN-Transformer encoders allow for the simultaneous extraction of features across varying scales and properties \cite{TransCloudSeg}. Furthermore, Guo et al. proposed the channel attention mechanisms that are frequently embedded within encoder blocks to recalibrate feature responses \cite{Guo}.

However, many of these methods rely on simple concatenation for fusion, potentially overlooking spatial-channel interactions and incurring high computational costs that hinder real-time deployment.

\subsection{Decoder for Ground-based Cloud Image Segmentation}
The primary function of the decoder is to progressively generate task-aligned outputs from the enriched feature vectors. A critical challenge here is recovering the spatial details lost during downsampling. Standard upsampling operations often result in blurred boundaries. To address this, Shi et al. have introduced bilinear upsampling and, more notably, content-aware feature reassembly operators. These advanced operators dynamically predict reassembly kernels based on local content, significantly reducing information loss and sharpening boundary delineation \cite{CloudU-Netv2, CloudFU-Net}. Moreover, to improve generalization across diverse atmospheric conditions, Li et al. proposed dynamic weight generation mechanisms in the final classification layers, allowing the model to adaptively adjust its decision boundaries based on aggregated encoder-decoder features \cite{DDUNet}. Effective fusion of high-level semantics and low-level details is essential for precise segmentation. Bilateral segregation and aggregation strategies have been proposed to leverage coarse segmentation priors to guide the refinement of fine-grained features \cite{BSANet}. Similarly, Feng et al. constructed multi-branch adaptive fusion units to intelligently merge features from different network depths, balancing local and global information \cite{MFAFNet}. To further strengthen boundary integrity, attention mechanisms are extensively applied in the decoder. Convolutional block attention modules are frequently integrated to enhance feature restoration during upsampling \cite{MA-SegCloud, MFAFNet}. For modeling complex, non-local relationships between disjoint cloud regions, Shi and Liu et al. have employed global graph reasoning frameworks. These methods project features into an interaction space to infer long-range dependencies before mapping them back to the pixel space \cite{TransCloudSeg, CloudSwinNet}. Additionally, token-based dual attention mechanisms have been utilized in skip connections to bridge the encoder and decoder, enhancing feature discriminability. Despite these innovations, the computational complexity of such attention mechanisms remains a concern for inference efficiency \cite{MA-SegCloud, DASUNet}.

Notwithstanding these innovations, these approaches fail to capture cross-hierarchical global feature correlations during upsampling, and the computational complexity of such attention mechanisms remains a concern for inference efficiency.
\subsection{Ground-based Cloud Image Dataset}
The proliferation of DL techniques is predicated on the availability of high-quality datasets captured by all-sky imagers. Several public benchmarks have been released to facilitate research.

Li et al. introduced the UTILITY dataset for analyzing statistical characteristics of clouds and sky elements \cite{dataset1}. Thirty-two images were manually converted into binary mask images with a resolution of 682 $\times$ 512 using a Voronoi polygon-based region generator. Dev et al. made publicly available a large-scale annotated database of sky/cloud images to the research community, named the Singapore whole sky imaging segmentation database (SWIMSEG) \cite{dataset8}. The SWIMSEG dataset comprises 1,013 images captured by the wide angle high-resolution sky imaging system (WAHRSIS), developed and deployed at Nanyang Technological University, Singapore, with each image having a resolution of 600 $\times$ 600 pixels \cite{dataset3}. Additionally, Dev et al. released the Singapore whole sky nighttime imaging segmentation database (SWINSEG) \cite{dataset2}, a dedicated dataset for nighttime cloud segmentation, consisting of 115 images at a resolution of 500 $\times$ 500 pixels. Fabel et al. collected data using an all-sky imager based on a commercial surveillance camera (Mobotix model Q25), assembling a total of 770 all-sky images (ASI) with a resolution of 512 $\times$ 512 pixels \cite{dataset4}. Building upon the SWIMSEG, Park et al. incorporated additional hazy and overcast images captured by a Waggle sensor node equipped with a camera, all of which have a resolution of 300 $\times$ 300 pixels \cite{dataset5}. Liu et al. published the Tianjin Normal University large-scale cloud detection database (TLCDD), which contains 5,000 cloud images. All images are stored in PNG format with a pixel resolution of 512 $\times$ 512 \cite{dataset6}. Kalisch et al. employed an all-sky imager developed by the Leibniz Institute of Marine Sciences (IFM-GEOMAR) at the University of Kiel. The system is equipped with a digital CCD camera featuring a fisheye lens, providing a 183° field of view. The sky images are stored in 30-bit color JPEG format with a maximum resolution of 3648 $\times$ 2736 pixels \cite{dataset7}. Shi et al. selected 300 images from the dataset released by \cite{dataset7}, cropped them to a resolution of 1704 $\times$ 1704, and publicly released the fine-grained segmentation dataset. By incorporating radiation characteristics, this dataset supports fine-grained segmentation into five categories: cumulonimbus and laminatus, cumulus, stratus, cirrus, and sky\cite{CloudFU-Net}.

While these datasets have significantly advanced the field to the development of the community, they frequently neglect the challenge of blurred cloud pixel boundaries around radiative sources, which arises due to solar-induced color distortion effects on cloud formations.
\begin{figure}[!t]
	\centering{\includegraphics[width=3.1in]{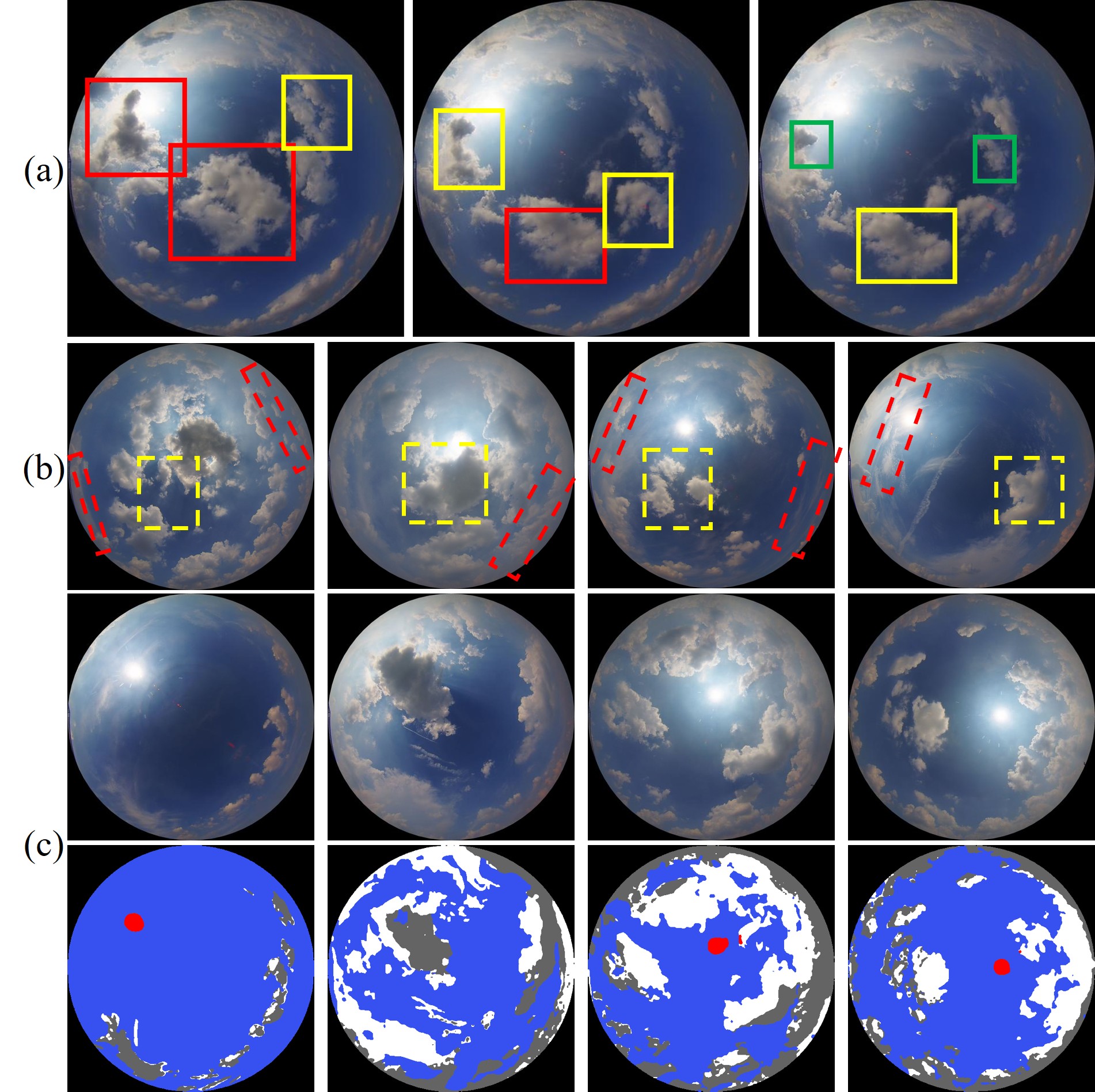}}
	\caption{(a) illustrates multi-scale clouds. The red, yellow, and green solid line rectangles represent large, medium, and small-scale cloud clusters, respectively. Across sequential frames, identical cloud formations exhibit varying scale variations. (b) illustrates the examples of elongated strip-like cloud bands near the horizon in CSRC. The yellow and red dashed rectangles indicate the zenith region (image center) and the horizon region (image edge), respectively. (c) illustrates the incomplete extraction of the boundary of clouds. Note that the results are from U-Net.}
	\label{fig_0}
\end{figure}
\begin{algorithm}[t!]
	\setlength{\intextsep}{0pt}
	\setlength{\abovedisplayskip}{0pt}
	\setlength{\belowdisplayskip}{0pt}
	\caption{Procedure of the MPCM-Net}
	\label{alg:mpcmnet}
	\SetKwInOut{Input}{Input}
	\SetKwInOut{Output}{Output}
	\SetKwRepeat{Repeat}{repeat}{until}
	
	\Input{Input image $X_{i} \in \mathbb{R}^{3 \times H \times W}$}
	
	\Output{Segmented image $Y_{i} \in \{0, 1, 2, 3\}^{H \times W}$ }
	\BlankLine
	
	\Repeat{convergence}{
		Let layer $i$ = 1, $loss$ = 0.0
		
		$L_{joint}=\alpha L_{focal}+\beta L_{dice}$ ($L$: loss function)
		
		\For{$i \leftarrow 1$ \KwTo $3$}{
			$f_{i} \leftarrow \text{MPC}_{i}$ \tcp*{MPC layer $i$}
		}
		
		
		$f_{4} \leftarrow \text{MPA}(f_{3})$ \tcp*{MPA layer}
		
		
		$u_{4} \leftarrow f_{4}$ 
		
		\For{$j \leftarrow 3$ \KwTo $1$}{
			$u_{j} \leftarrow \text{UP}(u_{j+1}, f_{j})$ \tcp*{Decoder layer $j$}
		}
		$f_{d} \leftarrow \text{Concat}(u_{2}, \text{UP}(u_{3}), \text{UP}(u_{4}))$
		
		$f_{s} \leftarrow \text{M2B}(f_{d})$ \tcp*{M2B block}
		
		$Y \leftarrow \text{Conv}_{1\times1}(\text{Concat}(f_{s}, u_{1}))$ 
		
		$\mathcal{L} \leftarrow \lambda_{1}L_{focal}+\lambda_{2}L_{dice}$  
		
		$\theta \leftarrow \theta - \nabla_\theta \mathcal{L}$ \tcp*{Parameter update}
	}
	\Return $Y_{i}$
\end{algorithm}
\vspace*{-10pt}

\begin{figure*}[!t]
	\centering{\includegraphics[width=6in]{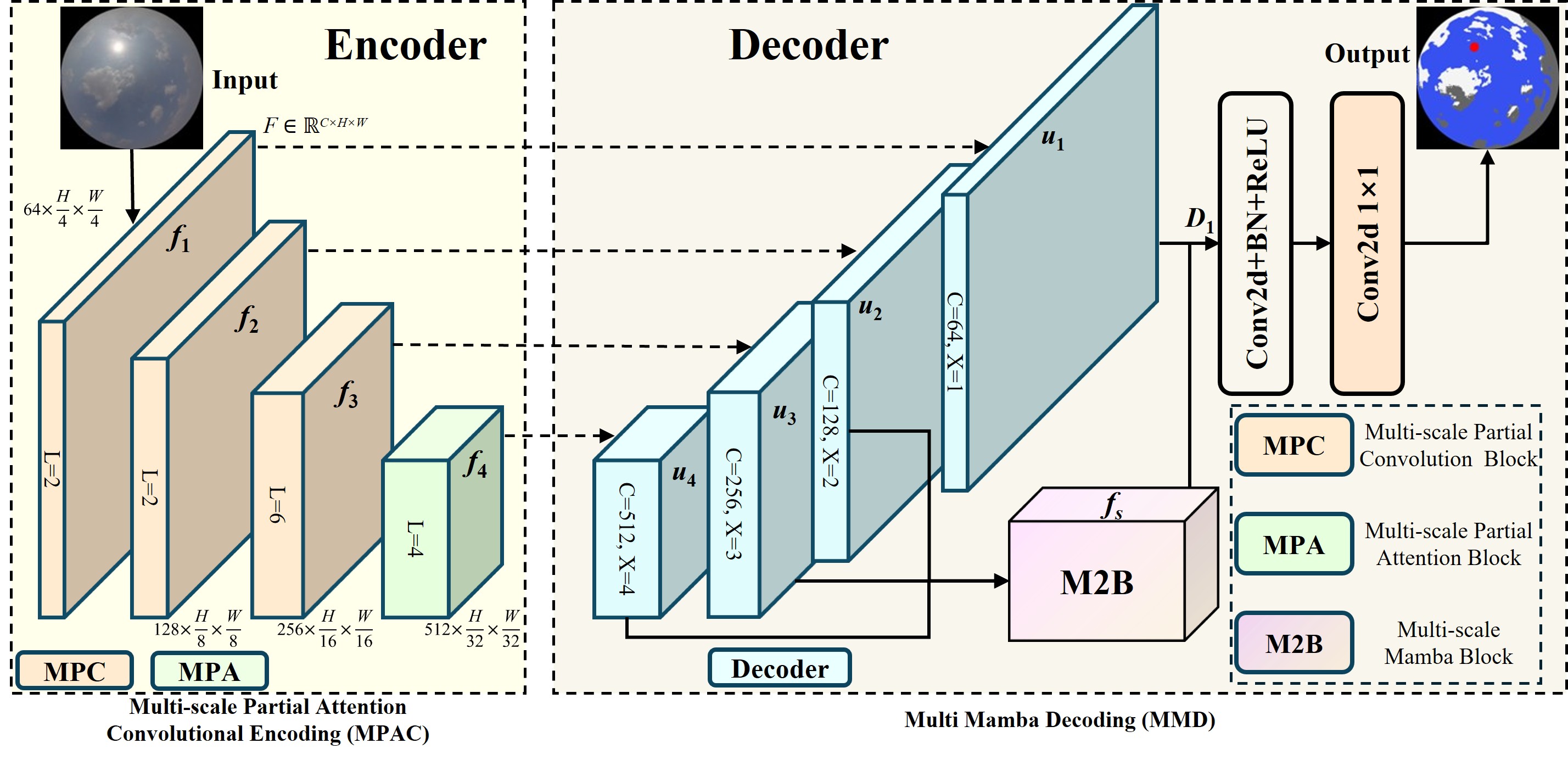}}
	\caption{The structure of the proposed MPCM-Net, which is composed of two parts: Multi-scale Partial Attention Convolution Encoding and Multi-scale Mamba Decoding. The MPAC comprises two key components: the MPC block and the MPA block. The MMD incorporates an M2B block.}
	\label{fig_1}
\end{figure*} 

\section{Proposed Method} \label{sec:method}
This section mathematically formalizes the ground-based cloud image segmentation task and subsequently delineates the architectural principles of the proposed MPCM-Net.

\subsection{Problem Description}
Formally, given an input image $X_{i}\in\mathbb{R}^{C \times H \times W}$, where $ C$, $H $, and $ W $ are channel, height, and width dimensions, respectively, the goal is to generate a corresponding segmentation mask $Y_{i} \in \{0, 1, 2, 3\}^{H \times W}$, where each pixel is classified as either cloud or non-cloud. Ground-based imagery inherently captures high-resolution cloud structures characterized by dynamic evolutionary patterns and pronounced multi-scale variations in both texture and morphology, as illustrated in Fig.~\ref{fig_0} (a). Furthermore, as depicted in Fig.~\ref{fig_0} (b), cloud boundaries, especially at different scales and near radiation sources, often manifest as translucent regions. These regions are spectrally distinct from the opaque cloud core yet lack sharp gradients against the sky background. These characteristics cause conventional convolution-based models to lose fine local features during encoding, leading to inaccurate boundary segmentation and further causing “ghosting effects” during cloud motion extrapolation tasks. Consequently, the primary challenge lies in designing a segmentation model that can effectively extract and integrate multi-scale contextual features while preserving fine boundary details. Motivated by this, we propose a novel network architecture that incorporates partial convolution and a Mamba-based decoder to enhance both segmentation accuracy and inference efficiency.

\subsection{Overview of Structure}
The MPCM-Net adopts a classic encoder-decoder paradigm (Fig.~\ref{fig_1}) designed to explicitly model multi-scale cloud features and resolve boundary ambiguity through a hierarchical feature interaction mechanism. The architecture comprises two primary subsystems: the MPAC encoder and the Multi-scale MMD module.
	
Algorithm~\ref{alg:mpcmnet} details the procedural execution of the network. In the encoding phase, the MPAC extracts robust contextual features via a synergistic integration of the MPC block and the MPA block. The MPC further decomposes into the ParCM and the ParSM (Fig.~\ref{fig_2}), ensuring efficient feature extraction across dimensions.

Subsequently, the decoder employs the MMD, which incorporates a M2B to reconstruct spatial details. A critical innovation within the M2B is the SSHD mechanism, which facilitates comprehensive feature learning across both global and local receptive fields by leveraging the linear complexity of State Space Models. The following subsections provide a granular explication of these components.

\subsection{Multi-scale Partial Attention Convolution Encoding}
Existing methodologies often extract multi-scale information from cloud clusters across consecutive frames, a process that frequently relies on standard dilated convolutions. This reliance typically neglects inter-channel feature interoperability and incurs significant latency penalties. To mitigate these limitations, the MPAC module introduces a partial convolution mechanism \cite{MALA} designed to enhance feature representation within distinct local regions.

\begin{figure*}[!t]
	\centering{\includegraphics[width=7in]{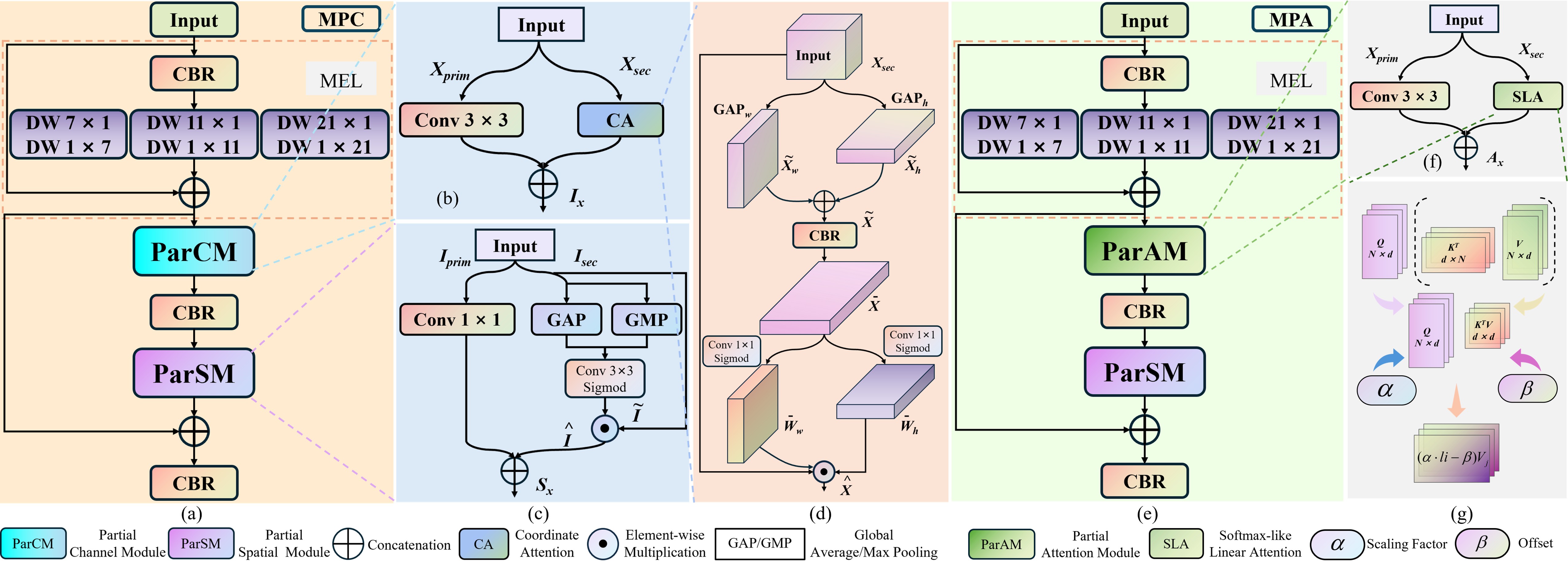}}
	\caption{The structure of (a) the MPC block and (e) the MPA block; (b), (c), and (g) are the structures of the ParCM, the ParSM, and the ParAM, respectively; (d) is the structure of the CA.}
	\label{fig_2}
\end{figure*}

As illustrated in the left panel of Fig. \ref{fig_1}, the encoder comprises four consecutive convolutional stages. The initial three stages utilize Multi-scale Partial Convolution (MPC) blocks, while the final stage employs the Multi-scale Partial Attention (MPA) block. Each stage is preceded by a Multi-scale Extraction Layer (MEL), as shown in Fig.\ref{fig_2} (a).
The MEL transforms input features $F \in \mathbb{R}^{C \times H \times W}$ through a CBR block comprising a $1 \times 1$ convolution, Batch Normalization (BN), and a Rectified Linear Unit (ReLU) activation. Subsequently, the features traverse three parallel strip convolutions with kernel scales of 7, 11, and 21, respectively. This kernel sizing strategy aligns with established protocols in segmentation literature \cite{SegNeXt, LG-Umer}. The encoder generates a hierarchical feature pyramid $f_{i}$ for $i \in \{1, 2, 3, 4\}$, where the resolution of $f_{i}$ is $H/2^{i} \times W/2^{i}$ with channel dimensions scaling as $64 \cdot 2^{i-1}$.

We incorporate the strip convolution mechanism to effectively capture the anisotropic characteristics of cloud formations\cite{Strip, ACNet}. Ground-based imagers project the hemispherical sky onto a 2D plane, causing cloud bands to appear as elongated “strips”, particularly near the horizon\cite{MFAFNet}. To provide an intuitive justification for the necessity of strip convolution, Fig.~\ref{fig_0} (b) illustrates the pronounced differences in cloud morphology between the zenith and horizon regions in CSRC. Near the image center, cloud formations preserve their complete geometric characteristics. Toward the image periphery, however, the mirroring-induced compression inherent to spherical projection deforms the clouds into banded, strip-like structures. Conventional square kernels ($k \times k$) introduce redundancy and background noise when processing such geometries. By decomposing a $k \times k$ convolution into a horizontal ($1 \times k$) and a vertical ($k \times 1$) strip convolution, the operation can be formulated as: $Y = f(X \cdot W_{1 \times k} + X \cdot W_{k \times 1})$, where $\cdot$ denotes the convolution operation. This decomposition reduces computational complexity from $O(k^2)$ to $O(2k)$, a critical optimization for enabling real-time inference in onsite PV power forecasting systems.

\subsubsection{Partial Channel Module}
To alleviate the computational redundancy inherent in multi-scale feature channels, we propose a partial channel mechanism integrated with coordinate attention. As depicted in Fig. \ref{fig_2} (b), given a feature map $ X \in \mathbb{R}^{C \times H \times W}$, input feature channels are partitioned based on a splitting ratio $r$ (empirically set to 0.25 based on the trade-off between FLOPs reduction and feature redundancy, consistent with FasterNet \cite{FasterNet}). The channel dimension is split into a primary part $C_{prim}$ and a secondary part $C_{sec}$ as follows: $C_{prim} = \lfloor C \cdot r \rfloor$ and $C_{sec} = C - C_{prim}$. Accordingly, the input has been split into two parts: $X_{prim} \in \mathbb{R}^{C_{prim} \times H \times W}$ and $X_{sec} \in \mathbb{R}^{C_{sec} \times H \times W}$. The subset $ X_{prim}$ is processed via a $3 \times 3$ convolution to aggregate global context, while the complementary sub-feature $ X_{sec}$ leverages a channel attention mechanism to model spatial interactions.

Unlike PartialNet \cite{PartialNet}, which targets general segmentation tasks, our design addresses the unique challenges of scale variations and stochastic motion. We therefore introduce coordinate attention (CA) instead of SE-Net\cite{SE-Net} to enhance channel-wise feature representation of cloud structures, as illustrated in Fig.~\ref{fig_2} (d).

Specifically, global average pooling is applied to the input $ X_{sec}$ along the horizontal and vertical axes, yielding coordinate vectors $ \tilde{X}_{w} \in \mathbb{R}^{C_{sec} \times H \times 1}$ and $ \tilde{X}_{h} \in \mathbb{R}^{C_{sec} \times 1 \times W}$. These vectors are concatenated to form a spatial descriptor $ \tilde{X}\in \mathbb{R}^{C_{sec} \times 1 \times (H+W)}$, which encodes directional information essential for tracking cloud motion. This descriptor undergoes a convolutional transformation to produce $\bar{X}\in \mathbb{R}^{C_{sec} \times 1 \times (H+W)}$, which is subsequently split into direction-specific tensors $ \bar{X}_{w} \in \mathbb{R}^{C_{sec} \times H \times 1}$ and $ \bar{X}_{h} \in \mathbb{R}^{C_{sec} \times 1 \times W}$. These tensors are processed via sigmoid activation to generate attention weights $ \bar{W}_{w} \in \mathbb{R}^{C_{sec} \times H \times 1}$ and {$ \bar{W}_{h} \in \mathbb{R}^{C_{sec} \times 1 \times W}$. This process injects directional priors into the feature map, allowing the model to focus on the fine-grained motion characteristics of cloud boundaries. The final output $\hat{X}$ is obtained via:
\begin{equation}
	\begin{aligned}
		\begin{split}
			\tilde{X} &= \left [ \tilde{X_{w} },\ \tilde{X_{h} }  \right ]\\ &=   \left [ G_{w}\left ( X_{sec} \right ),\ G_{h}\left ( X_{sec} \right )   \right ]
		\end{split}
	\end{aligned}
	\label{equ_1}
\end{equation}
\begin{equation}
	\bar{X} = \mathrm {Conv} \left ( \tilde{X}  \right )
\end{equation}
\begin{equation}
	\bar{W}_{w}= \sigma \left ( \mathrm {conv}\left ( \bar{X} _w  \right )  \right ),
	\bar{W}_{h}= \sigma \left ( \mathrm {conv}\left ( \bar{X} _h  \right )  \right )
	\label{equ_1.1}
\end{equation}
\begin{align}
	\hat{X}= X_{2}\odot \bar{W}_{w} \odot \bar{W}_{h}
	\label{equ_2}
\end{align}\noindent where $ \left [ \cdot , \cdot  \right ] $ denotes the concatenation, $G_{w} \left( \cdot  \right)$ and $G_{h} \left( \cdot  \right)$ denote the global average pooling along the width dimension and height dimension, respectively. $\mathrm {Conv} \left( \cdot  \right)$ denotes the convolutional operation, $\mathrm {conv} \left( \cdot  \right)$ denotes the 1 $\times$ 1 convolution. $\sigma \left( \cdot  \right)$ denotes the sigmoid function and $\odot $ denotes the element-wise multiplication. The output of the ParCM is $I_{x}$:
\begin{align}
	I_{x}=\left[ \mathrm {conv_{3\times 3}}\left( X_{prim}\right), \hat{X} \right]
	\label{equ_2.1}
\end{align}

\subsubsection{Partial Spatial Module}
While downsampling operations across successive encoder layers inevitably degrade spatial location information in feature maps, this property paradoxically mitigates boundary ambiguity. As illustrated in Fig.~\ref{fig_2} (c), firstly, the input features are divided into two sub-features, $I_{prim} \in \mathbb{R}^{C_{prim} \times H \times W}$ and $I_{sec} \in \mathbb{R}^{C_{sec} \times H \times W}$, by a ratio $r$. Then we adapt the ParCM paradigm by integrating partial channel information with an attention mechanism and a 1 $\times$ 1 convolution for effective channel mixing. Unlike conventional spatial attention, relying solely on pointwise convolution, our approach employs a spatial selectivity mechanism to capture precise positional information. This mechanism utilizes spatial feature descriptors to focus on the most relevant spatial regions adaptively. Specifically, given input feature map $I_{sec}$, we apply global max pooling and global average pooling to derive two complementary feature maps enriched with detailed spatial information. These maps are concatenated along the channel dimension to enhance global context. A convolutional operation reduces the fused feature dimensionality. A sigmoid activation produces the spatial attention descriptor $\tilde{I}$ for weighting multi-scale features to obtain the final output $\hat{I}$. The specific formula for this process is as follows:
\begin{align}
	\tilde{I}&= \sigma \left ( \mathrm{conv}\left( \left [ \mathrm{GMP}\left ( I_{sec} \right ), \mathrm{GAP}\left ( I_{sec} \right )  \right ] \right )  \right ) \\
	\hat{I}& =\tilde{I} \ast I_{sec}
	\label{equ_3}
\end{align}\noindent where $\mathrm{GAP}\left( \cdot  \right)$ and $\mathrm{GMP}\left( \cdot  \right)$ are the global average pooling and global max pooling, respectively. The output of the ParSM is $S_{x}$:
\begin{align}
	S_{x}=\left[ \mathrm {conv}\left( I_{prim}\right), \hat{I} \right]
	\label{equ_3.1}
\end{align}

\subsubsection{Partial Attention Module}
In the final layer of MPAC, to adaptively capture global semantic dependencies, we propose ParAM that adopts a “split-and-process” strategy consistent with the design philosophy of ParCM and ParSM.

As illustrated in Fig.~\ref{fig_2} (f), we first split the input feature map. The sub-feature $X_{sec}$ is processed by our proposed Softmax-like Linear Attention (SLA) (The subsequent text will provide a comprehensive description) to focus on extracting long-range contextual dependencies. Finally, the outputs from two branches are concatenated.

An appropriately designed attention mechanism in ParAM significantly enhances segmentation precision for multi-scale cloud clusters. ViTs suffer from quadratic computational complexity $O$ ($N^{2}$) relative to input sequence length $N$\cite{ViT}. Linear attention addresses this limitation by reformulating the computation order of Query-Key-Value operations and eliminating softmax, thereby reducing complexity to $O$($N$) \cite{LinearA}. However, this efficiency gain comes at the expense of substantial performance degradation. Inspired by MALA \cite{MALA}, we introduce an amplitude-aware mechanism into the computation logic of the linear attention, proposing a novel Softmax-like Linear Attention mechanism. SLA preserves the amplitude propagation capability characteristic of softmax attention.

The performance degradation in linear attention mechanisms stems from their inability to propagate Q-amplitude values. Consequently, attention scores across different features remain confined to an identical distribution, failing to dynamically focus on salient regions in response to variations in Q-values. This mechanism utilizes the kernel function $\phi$ (.) to approximate the similarity function and maps $Q$ and $K$ to positive real numbers. The core innovation of SLA addresses this limitation by introducing an offset term to amplify weights associated with higher original attention scores, thereby enhancing local feature concentration. Specifically,  as illustrated in Fig.~\ref{fig_2} (g), we reformulate the similarity computation by incorporating a scaling factor $\alpha$ and offset $\beta$ into the original linear attention framework, formulating a novel attention metric as follows:
\begin{figure}[!t]
	\centering{\includegraphics[width=3.5in]{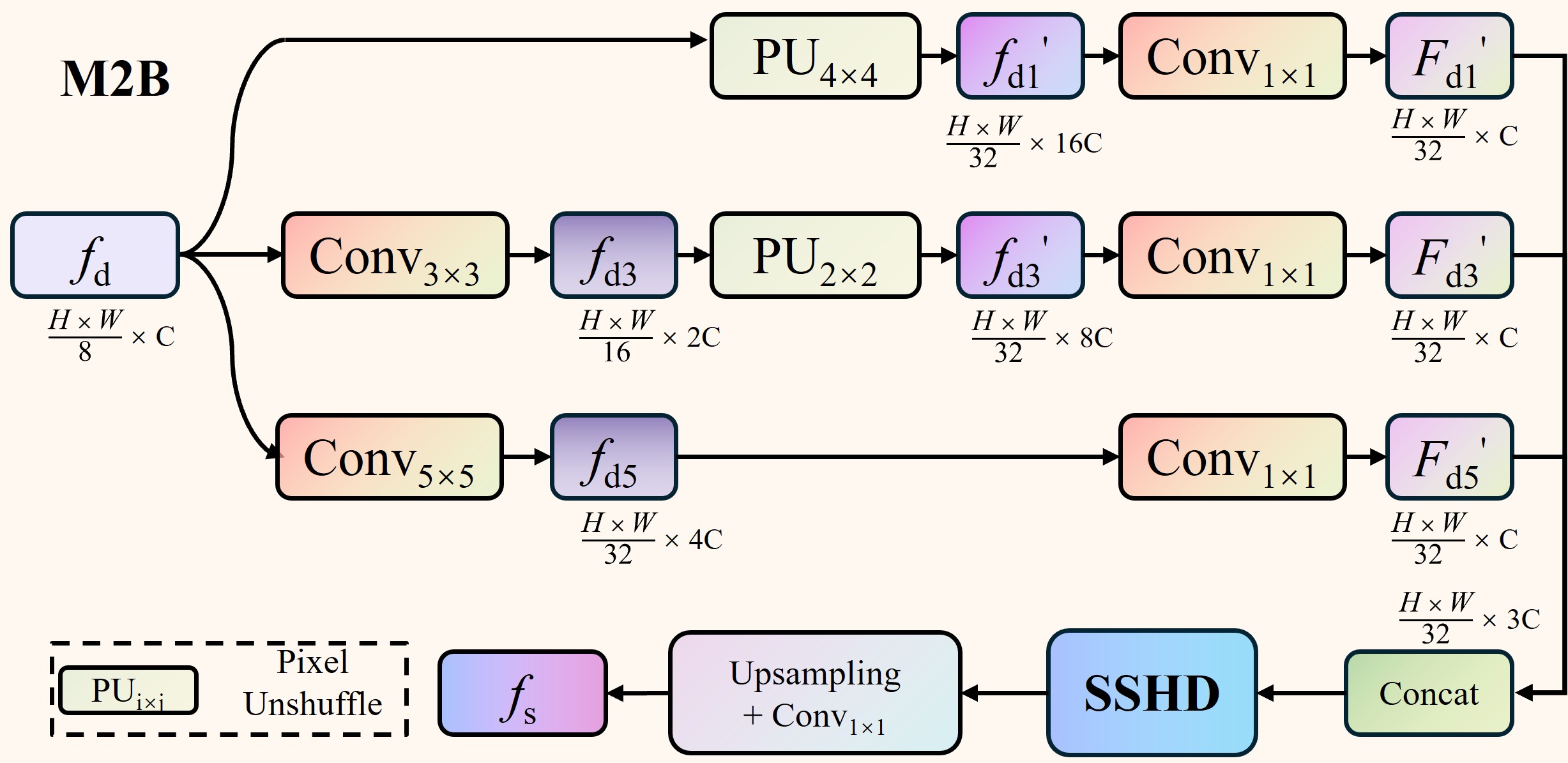}}
	\caption{The overall architecture of the M2B module comprises three parallel branches that extract multi-scale features with varying receptive fields. These features are subsequently fused and processed through the SSHD to generate the final multi-scale global feature representation, denoted as $f_{s}$.}
	\label{fig_3}
\end{figure}
\begin{equation}
	Q=xW_{Q},K=xW_{K},V=xW_{V}
\end{equation}
\begin{equation}
	\begin{aligned}
		\begin{split}
			\mathrm{Att}_{i}&=\sum_{j=1}^{N} \frac{\phi\left(Q_{i}\right) \phi\left(K_{j}\right)^{T}}{\sum_{m=1}^{N} \phi\left(Q_{i}\right) \phi\left( K_{m} \right)^{T}} V_{j}\\
			&=\sum_{j=1}^{N} \frac{l_{ij}}{\sum_{m=1}^{N} l_{im}} V_{j}
		\end{split}
	\end{aligned}
	\label{equ_4}
\end{equation}
\begin{equation}
	\begin{aligned}
		\begin{split}
			\mathrm{SLA}_{i}&=\sum_{j=1}^{N} \left[\frac{l_{ij}}{\sum_{m=1}^{N} l_{im}} \left(1+\sum_{m=1}^{N} l_{im}\right)-\frac{\sum_{m=1}^{N} l_{im}}{N} \right]V_{j}\\
			&=\sum_{j=1}^{N} \left[\left(1+\frac{1}{\sum_{m=1}^{N} l_{im}} \right)l_{ij}-\frac{\sum_{m=1}^{N} l_{im}}{N} \right]V_{j}\\
			&=\sum_{j=1}^{N} \left( \alpha \cdot l_{ij}-\beta  \right)V_{j}
		\end{split}
	\end{aligned}
	\label{equ_5}
\end{equation}\noindent where $\mathrm{Att}_{i}$ denotes the original linear attention with an input of $N$ tokens which represented as $ x\in \mathbb{R}^{N\times C} $. $l_{\left( xy \right )}$ denotes $ \phi\left(Q_{x}\right) \phi\left(K_{y}\right)^{T}$. $\phi(\cdot)$ denotes the kernel feature map function. We use $\phi(x) = \mathrm {ELU} \left(x \right) + 1$, where $\mathrm {ELU}$ denotes the Exponential Linear Unit: $\mathrm {ELU}\left(x \right) = e^x - 1$ for $x \le 0$, $x$ for $x > 0$. The $+1$ shift ensures that all kernel values are strictly positive, which is a mathematical prerequisite for the validity of the subsequent attention probability mass function \cite{LinearA}. $\alpha =\left(1+1/\sum_{m=1}^{N}l_{im} \right)$ denotes the scale factor and $\beta=1/N\cdot \sum_{m=1}^{N} l_{im}$ denotes the offset term. 

Given $Q_{i}$, the ratio of its attention scores toward two distinct keys $K_{n}$ and $K_{t}$ can be expressed as follows:

\begin{equation}
	\frac{\alpha l_{in}-\beta }{\alpha l_{it}-\beta } =w
	\label{equ_6}
\end{equation}
Assuming $Q$ assigns a higher attention score to $K_{n}$, when the magnitude of $Q$ scales by a factor $s>1$, the updated values of $\alpha$ and $\beta$ can be derived as:
\begin{equation}
	\begin{aligned}
		\begin{split}
			\alpha _{up}&=1+\frac{1}{s\sum_{m=1}^{N} l_{im}}\\
			&=1+\frac{1}{s} \left(\alpha -1\right) =\frac{s+\alpha -1}{s}
		\end{split}
	\end{aligned}
	\label{equ_7.1}
\end{equation}
\begin{equation}
	\begin{aligned}
		\begin{split}
			\beta _{up}=\frac{s\sum_{m=1}^{N} l_{im}}{N}=s\beta
		\end{split}
	\end{aligned}
	\label{equ_7.2}
\end{equation}
Consequently, $w_{up}$ updates to:
\begin{equation}
	\begin{aligned}
		\begin{split}
			w_{up}&=\frac{\alpha_{up} l_{in}-\beta_{up} }{\alpha_{up} l_{it}-\beta_{up} }\\
			&=\frac{\left( s+\alpha -1\right )l_{in}-s\beta}{\left(s+\alpha -1\right )l_{it}-s\beta}\\
			&=\frac{\alpha l_{in}-\frac{s\alpha }{s+\alpha -1}\beta }{\alpha l_{it}-\frac{s\alpha }{s+\alpha -1} \beta }
		\end{split}
	\end{aligned}
	\label{equ_8}
\end{equation}
Since $s > 1$, $\alpha > 1$, and $\frac{s\alpha }{s+\alpha -1} > 1$, it can be proven that $w_{up} > w$.

\noindent\textbf{Proof of} $w_{up} > w$.

We define $A=\alpha l_{in}$, $B=\alpha l_{it}$ and $k = \frac{s\alpha}{s+\alpha-1}$, where $s\alpha - (s + \alpha - 1) = (s-1)(\alpha-1) > 0$. We obtain $s\alpha > s + \alpha - 1$, so $k > 1$. We assume that $Q_{i}$ assigns a higher score to $K_{n}$, thus $A>B$. Consider the function $f(x) = \frac{A-x}{B-x}$. Its derivative with respect to $x$ is: $f'(x) = \frac{A-B}{(B-x)^2}$. Since $A > B$, $f'(x) > 0$, implying $f(x)$ is strictly increasing. Since $k\beta > \beta$, it follows that $f(k\beta) > f(\beta)$, which proves $w_{up} > w$.

This demonstrates that SLA allocates greater attention to features originally receiving higher scores while reducing attention to features with lower scores, thereby successfully emulating the strong representational capacity of standard softmax attention.
\begin{figure*}[!t]
	\centering
	\includegraphics[width=6.5in]{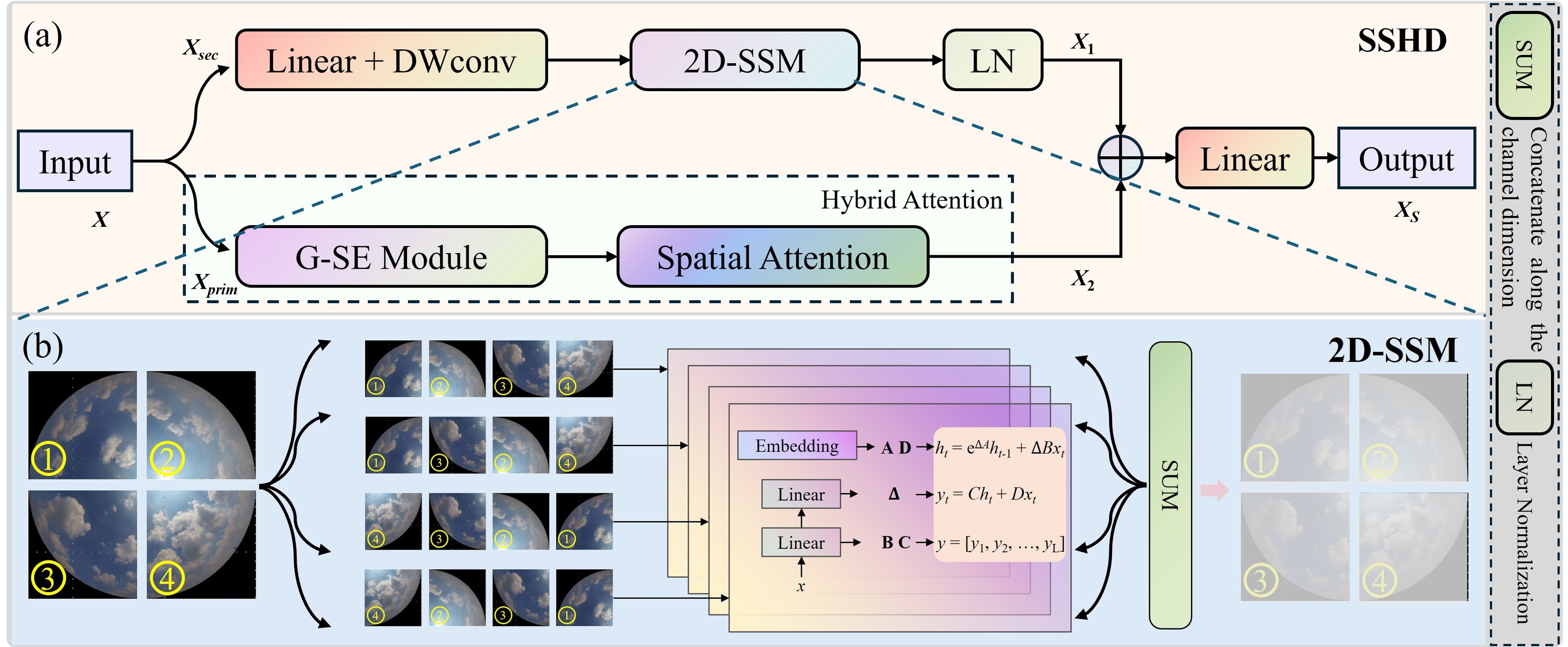}
	\caption{(a) The structure of the proposed SSHD is composed of two parts: $X_{sec}$ and $X_{prim}$. The $X_{sec}$ passes through the 2D-SSM, the $X_{prim}$ passes through the HA module, and obtains the SSHD output $X_{s}$. (b) The specific structure of 2D-SSM. The 2D feature map is first flattened into a 1D sequence, then the orientation-specific information is captured via a selective state space model. Finally, the scanned sequences are merged to reconstruct the original 2D feature. The right side of the figure shows the notes of some modules.}
	\label{fig_4}
\end{figure*}

\subsection{Multi Mamba Decoding}
Establishing global dependencies over long-range features enables the extraction of comprehensive cloud cluster information. Existing computationally expensive decoders lack global feature correlations between local information across hierarchical levels. In fact, Mamba \cite{Mamba} has been shown to establish global context. Nevertheless, its compressed fixed-dimensional state representation can lead to partial loss of local information.

To this end, we propose a decoder architecture incorporating a Multi-scale Mamba Block. Specifically, we design a multi-scale feature enhancement module based on the Mamba architecture within the decoder by introducing a spatial-semantic hybrid domain to process feature maps at varying granularities. The module achieves dynamic and adaptive extraction of multi-scale semantic information. Collectively, our decoder performs deep feature aggregation both spatially and across scales, significantly mitigating information loss caused by translucent cloud boundaries.

The proposed decoder, illustrated in the right side of Fig.~\ref{fig_1}, comprises four hierarchical levels. The final feature map $f_{4}$ of the encoder is upsampled to match the spatial resolution of the preceding level and concatenated with the encoder output feature map $f_{3}$, yielding feature map $u_{3}$. This process repeats recursively for each subsequent decoder layer, ultimately generating fused features $u_{i}\left( i\in \left \{ 1, 2, 3, 4 \right \}  \right )$, where $u_{2}$ possesses dimensions $H/8\times W/8$. Critically, the upsampling operation consists of two convolutional layers with 3 $\times$ 3 kernels followed by bilinear interpolation with a scale factor of 2. Subsequently, features $u_{3}$ and $u_{4}$ are upsampled to match the spatial resolution of $u_{2}$. These three-level features are then concatenated and fused, yielding a feature map $f_{d}$. This fused feature is then processed by the M2B to generate a feature map $f_{s}$, enriched with multi-scale contextual information. Finally, $f_{s}$ is upsampled and concatenated with $u_{1}$ to generate the final output of the proposed MPCM-Net.

\subsubsection{Multi-scale Mamba Block}
As illustrated in Fig.~\ref{fig_3}, to achieve multi-scale processing within the Mamba framework, we first employ large-kernel receptive field aggregation on $f_{d}$ to extract features at varying scales. Specifically, feature map $f_{d}$ is processed with a 3 $\times$ 3 convolution (stride = 2), aggregating neighborhood features within a 9 $-$ pixel receptive field, yielding $f_{d3}$ with dimensions $H /16\times W/16\times 2C$. Feature map $f_{d}$ is processed with a 5 $\times$ 5 convolution (stride = 4), aggregating neighborhood features within a 25 $-$ pixel receptive field, yielding $f_{d5}$ with dimensions $H/32\times W/32\times 4C$. The multi-scale feature maps are subjected to a pixel unshuffle procedure in order to allow the Mamba structure to progressively scan the channel dimensions of $f_{d}$, $f_{d3}$, and $f_{d5}$. This ensures uniform spatial dimensions across all three features while minimizing information loss inherent in downsampling. The core idea involves spatially rearranging $f_{d}$ and $f_{d3}$: non-overlapping spatial blocks are shifted into the channel dimension, preserving the spatial integrity of the original resolution while matching the reduced dimensions of $f_{d5}$, resulting in transformed features $f_{d1}^{'}$ and $f_{d3}^{'}$. The following equations formally describe this transformation:
\begin{align}
	f_{d1}^{'}&=\mathrm {PU}_{4\times4} \left(f_{d}\right)\\
	f_{d3}^{'}&=\mathrm {PU}_{2\times2} \left(f_{d3}\right)
	\label{equ}
\end{align}\noindent where $\mathrm {PU}_{i\times j} \left( \cdot  \right)$ denotes a pixel unshuffle operation using non-overlapping $i \times j$ image patches. Subsequently, the three feature maps are projected to a uniform channel dimension via 1 $\times$ 1 convolutions and concatenated along the channel axis. This concatenated feature is then fed into the proposed SSHD to enhance local and global features. The resulting feature map possesses $H/32\times W/32 \times3C$, where the three segments along the channel dimension correspond to the input features $F_{d1}^{'}$, $F_{d3}^{'}$, and $F_{d4}^{'}$, respectively. To maintain consistent spatial resolution throughout the decoder, the hybrid features output by SSHD are upsampled to match the spatial scale of $u_{2}$ via bilinear interpolation. Finally, a 1 $\times$ 1 convolution reduces dimensionality, yielding the multi-scale global feature $f_{s}$, which integrates local feature relationships. 

\subsubsection{Spatial-Semantic Hybrid Domain}
Previous methods present significant challenges for real-time inference requirements in ground-based cloud analysis. Motivated by the success of Mamba, we propose a SSHD module based on the Mamba architecture to enhance global feature representation.

The proposed SSHD module, illustrated in Fig.~\ref{fig_4} (a), processes input features similarly to the MPAC within the decoder. First, input feature $X \in \mathbb{R}^{C \times H \times W}$ undergoes channel partitioning with ratio $r$. The partitioned component $X_{sec}$ is then processed sequentially through a linear layer and a depth-wise convolutional (DW-Conv) layer to enhance feature representation. Subsequently, 2D State Space Model (2D-SSM) \cite{Vmamba} captures global contextual information incorporating local features, followed by layer normalization to genearte $X_{1}$. To extract fine-grained information, a parallel branch $X_{prim}$ introduces Hybrid Attention (HA), integrating channel and spatial attention mechanisms, for feature refinement to generate $X_{2}$. Finally, features from both branches are concatenated along the channel dimension and processed through a linear layer, yielding the SSHD output $X_{s}$. The complete workflow is formally described by the following equations:
\begin{align}
	X_{1}&=\mathrm {LN} \left( \mathrm {SSM} \left( \mathrm {DW}\left( \mathrm {Linear} \left(X_{sec}\right) \right) \right) \right)\\
	X_{2}&=\mathrm {HA}\left(X_{prim}\right)\\
	X_{s}& = \mathrm {Linear} \left ( \left[ X_{1},  X_{2} \right ] \right )
	\label{equ_10}
\end{align}\noindent where $\mathrm {LN}(\cdot)$ denotes Layer Normalization, and $\mathrm{Linear}(\cdot)$ denotes a fully connected projection layer.

Inspired by \cite{Vmamba}, we introduce a 2D selective scanning module. As illustrated in Fig.~\ref{fig_4} (b), the 2D feature map $X_{sec}$ is first flattened into a 1D sequence, then selectively scanned along four spatial directions: top-left to bottom-right, bottom-right to top-left, top-right to bottom-left, and bottom-left to top-right. During each directional scan, the selective state space model efficiently captures orientation-specific global contextual information and establishes long-range dependencies. Finally, the scanned sequences from all directions are merged to reconstruct the original 2D feature representation.

\begin{figure}[!t]
	\centering{\includegraphics[width=3.5in]{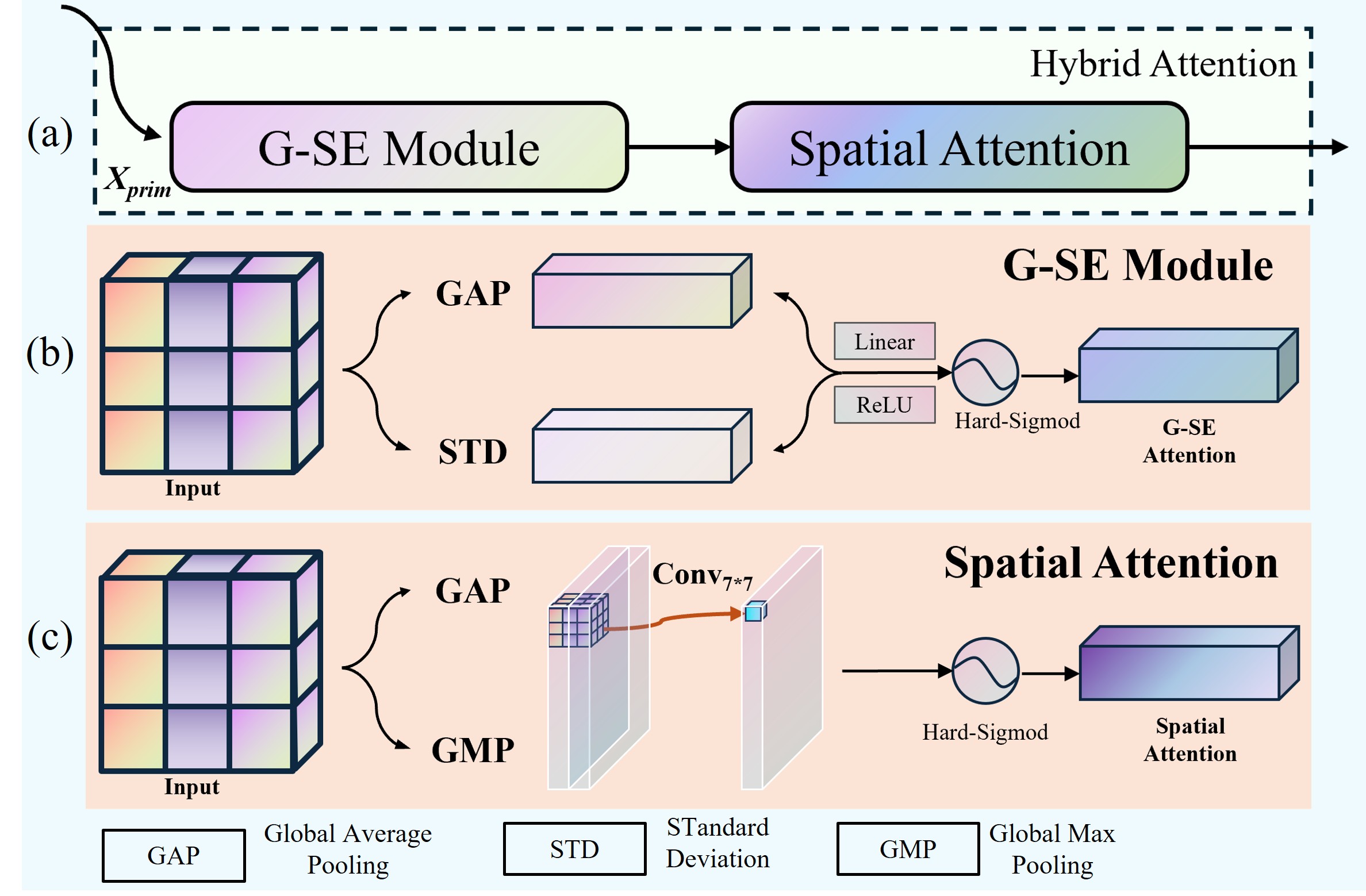}}
	\caption{(a) The overall structure of the proposed HA. (b) The proposed Gaussian-SE module. (c) The proposed spatial attention module. The lower side of the figure shows the notes of some modules.}
	\label{fig_5}
\end{figure}

We propose HA to enhance the interaction between spatial and semantic information in the MMD, as illustrated in Fig. ~\ref{fig_5}. Specifically, the feature subset $X_{prim}$ first undergoes channel attention. Since feature maps exhibit approximately Gaussian distributions during training, whereas standard Squeeze-and-Excitation blocks utilize only mean statistics \cite{SE-Net}, we introduce an enhanced Gaussian-SE module that compresses global spatial information using both mean and variance to model Gaussian statistics for channel-wise recalibration. Subsequently, spatial attention is implemented via point-wise convolution, compressing global channel information into a single-channel tensor. These tensors are processed through a Hard-Sigmoid activation function to generate the spatial attention map for feature weighting. The complete workflow is formally described by the following equations:
\begin{align}
	\mathrm {HA}&=\mathrm {S_{A}} \left( \mathrm{G_{SE}} \left(X_{prim}\right) \right) \\
	\mathrm{G_{SE}}&=\sigma_{H} \left( \mathrm{ReLU}(\mathrm{Linear}(\mathrm{GAP}\left ( X_{prim} \right ),\mathrm{STD}\left ( X_{prim} \right ))) \right ) \\
	\mathrm {S_{A}}& = \sigma_{H} \left( \mathrm {Conv_{7\times 7}} \left( \left [ \mathrm{GAP}\left ( X_{in} \right ),\mathrm{GMP}\left ( X_{in} \right )   \right ] \right) \right )
	\label{equ_11}
\end{align}\noindent where $\mathrm {S_{A}} \left( \cdot  \right)$ denotes spatial attention, $\mathrm{G_{SE}}$ denotes Gaussian-SE module, $\mathrm{STD}\left( \cdot  \right)$ denotes the standard deviation, $X_{in}$ denotes the input of SA, and $\sigma_{H}\left( \cdot  \right)$ is the Hard-Sigmoid activation function.

Finally, $f_{s}$ is upsampled and concatenated with $u_{1}$ to generate the resulting feature map $D_{1}$.
The $D_{1}$ undergoes a standard convolution to restore the input resolution, followed by a 1 $\times$ 1 convolution for channel adjustment prior to final prediction.

\subsection{Loss Function}
Ground-based cloud segmentation is intrinsically a binary pixel-level classification task characterized by severe class imbalance. Cloud formations often exhibit significant scale variations and low contrast against the sky background. To accelerate convergence and mitigate bias toward the background class, we employ a compound objective function:
\begin{equation}
	L_{joint}=\lambda_{1} L_{focal}+\lambda_{2} L_{dice}
	\label{equ_11}
\end{equation}\noindent where $ L_{focal}$ denotes the focal loss and $ L_{ dice }$ denotes the dice loss. $\lambda_{1}$ and $\lambda_{2}$ are both hyperparameters, set to 0.6 and 0.4, respectively.

The Focal Loss addresses the limitations of cross-entropy by introducing a modulating factor that down-weights easy negatives, forcing the model to focus on hard-to-classify boundary pixels. Complementarily, the Dice Loss optimizes the Intersection-over-Union directly, enhancing the model's sensitivity to the structural integrity of cloud masses.
\begin{equation}
	\begin{aligned}
		\begin{split}
			L_{focal}=-\frac{1}{M} \sum_{i=1}^{M} ( \gamma ( 1-p_{i} ) ^\delta \times p_{i} log ( p_{i} )+  \\
			( 1- \gamma ) p_{i}^\delta \times ( 1- g_{i} ) log ( 1- p_{i} ) )
		\end{split}
	\end{aligned}
	\label{equ_12}
\end{equation}
\begin{equation}
	 L_{ dice}=1-\frac{2\times \sum_{i=1}^{M} p_{i} g_{i} + \varepsilon  }{\sum_{i=1}^{M} (p_{i}^2+g_{i}^2 ) + \varepsilon    }
	\label{equ_13}
\end{equation}\noindent where $ p_{i}$ and $ g_{i}$ denote the prediction and ground truth of sample $i$, respectively. $M$ denotes the number of pixels used in the loss computation, and the index $i$ refers to pixels. $\gamma $ and $\delta $ denote the weight of the positive and negative samples. They are hyperparameters and the values are set to 0.25 and 0.2, respectively, as same as \cite{LG-Umer}. $\varepsilon $ denotes the smoothing coefficient, which is set to $ 10^{-5} $ as a hyperparameter to prevent the denominator in equation (\ref{equ_13}) from being zero. 

\section{Experimental Results and Discussions}\label{sec:exp}
This section presents comprehensive experiments validating the performance of the proposed MPCM-Net. We first introduce the constructed dataset, followed by detailed specifications of the experimental setup and evaluation metrics. Subsequent analysis evaluates MPCM-Net's performance through both qualitative and quantitative assessments. Finally, ablation studies demonstrate the efficacy of key proposed components. 

\begin{table}[!t]
	\centering
	\caption{Cloud Classification Standard for CSRC Dataset.}
	{\small
		\begin{tabular}{m{2cm}<{\centering}m{5cm}<{\centering}m{0.7cm}}
			\toprule
			Cloud class & Shape and characteristic & Color \\
			\midrule
			White cloud & \makecell{White transparent clouds, \\ such as cirrus and cirrocumulus, \\ are pure white and translucent \\ or appear as white scaly flakes.} & White \\
			Gray cloud & \makecell{Gray clouds, \\ such as stratus and altostratus, \\ are gray and foggy with a dark base.} & Gray \\
			Sun & Radiation source & Red \\
			Background & There are no clouds in the sky & Blue \\
			\bottomrule
	\end{tabular}}
	\label{Table0}
\end{table}
\subsection{Dataset} \label{sec:data}
A primary impediment in ground-based cloud image segmentation is the scarcity of high-quality, large-scale, and clear-label public datasets. The performance of DL models fundamentally depends on the diversity and precision of the training data. However, existing benchmarks primarily focus on binary classification of clouds versus sky background, neglecting important characteristics such as cloud scale, morphology, and color features. Recently, Shi et al. released a fine-grained ground-based cloud image segmentation dataset that incorporates radiation characteristics to classify images into five categories: cumulonimbus and laminatus, cumulus, stratus, cirrus, and sky, laying a foundation for fine-grained segmentation and PV power forecasting \cite{CloudFU-Net}. Nevertheless, since ground-based cloud segmentation is inherently a pixel-wise classification task, this dataset overlooks the influence of color attributes caused by radiative properties, as well as the impact of color variations within differently shaped clouds on segmentation accuracy. For example, in this dataset, both nimbostratus and cumulonimbus clouds with over 90$\%$ radiation attenuation rate and stratus clouds with approximately 50$\%$ attenuation rate exhibit gray colors. Similarly, cumulus clouds (with $>$ 90$\%$ attenuation rate) and cirrus clouds ($>$ 10$\%$) both appear white. The presence of similar color systems across different fine-grained categories hinders segmentation accuracy in pixel-level classification tasks. Moreover, this dataset is derived from the publicly available dataset by Ref\cite{dataset8} and contains only 300 images. Additionally, significant technical barriers remain in effectively leveraging off-site ground-based cloud image segmentation results for PV power prediction.

\begin{figure}[!t]
	\centering{\includegraphics[width=3.5in]{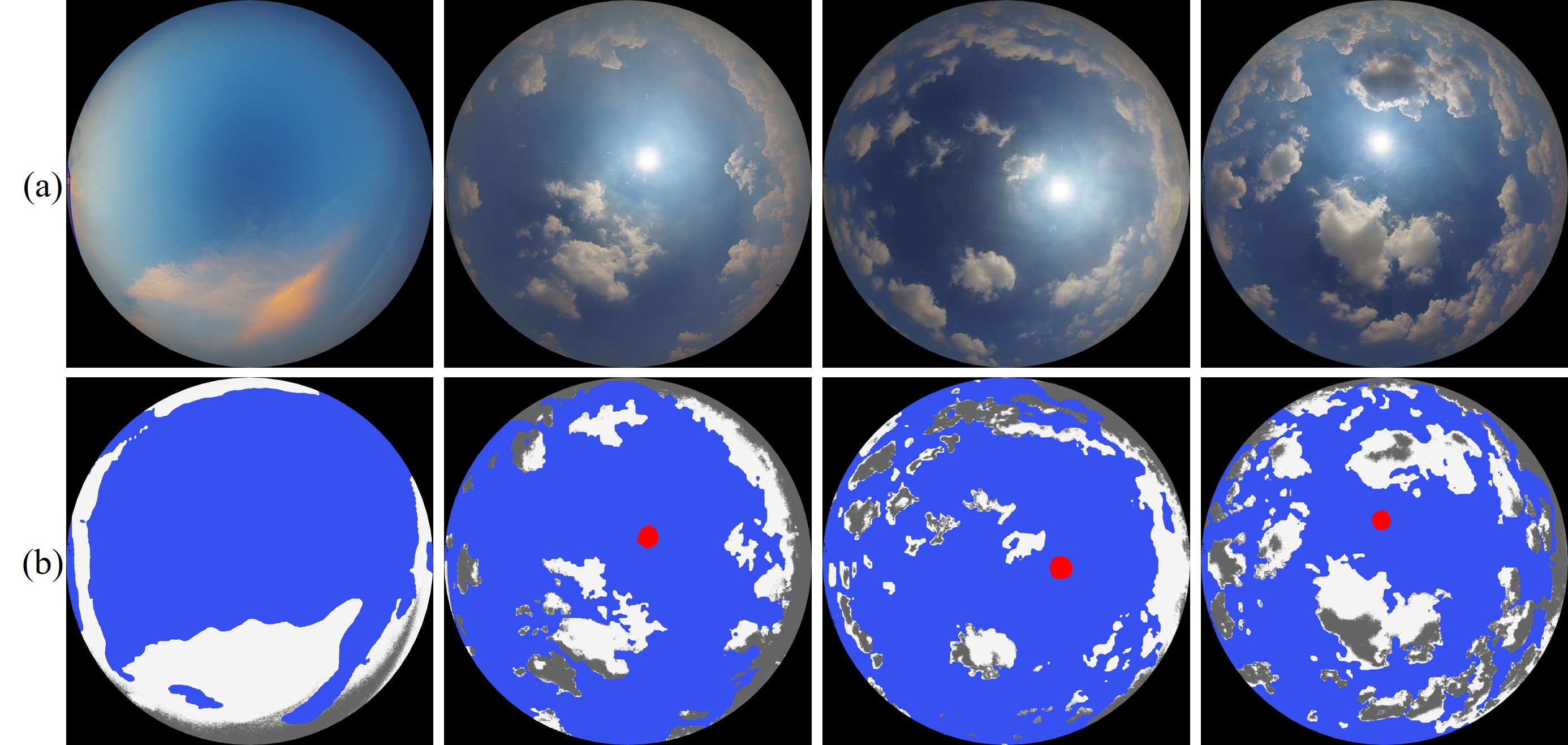}}
	\caption{(a) and (b) are the original image and ground truth from the CSRC dataset, respectively. The dataset contains the sun, white clouds, gray clouds, and background, corresponding to red, white, gray, and blue, respectively, and the images contain clouds of various scales.}
	\label{fig_6}
\end{figure}

To bridge this gap and provide the research community with a more challenging and practical benchmark, we developed and publicly released a new dataset CSRC, which constitutes a core contribution of this paper. The solar radiation effect on clouds is a critical factor in the accuracy of PV power forecasting. International cloud classification standards are primarily based on the brightness, color, size, and altitude of clouds. Among these, color variations induced by radiative sources play a significant role in the precision of pixel-level classification in ground-based cloud imagery. In this study, we focus particularly on cloud scale and color attributes, and introduce fine-grained segmentation that incorporates radiative influence, re-categorizing clouds into three classes (see Table~\ref{Table0} for details). White and gray clouds are common color patterns in the sky, often covering extensive areas, and are represented by white and gray labels, respectively. Pixels within the circumsolar region often appear nearly pure white due to high radiation, with red and blue channel values converging, leading these regions to be classified as overcast. Therefore, it is essential to incorporate fine-grained solar radiation segmentation \cite{dataset7}, which is indicated in red. Black cloud systems, typically associated with heavy rain or hail, result in full sky coverage and are detrimental to both ground-based cloud image segmentation and PV power prediction tasks. Such black clouds are ignored from our dataset. Cloud-free sky regions are labeled in blue as background.
Table~\ref{Table0_1} presents a comparative analysis of the proposed CSRC dataset and existing public benchmarks, evaluating performance across critical dimensions including fine-grained categorization support, radiation source diversity, color attribute availability, spatial resolution, and sample size. Regarding data completeness, it should be noted that the specific sample size is not explicitly documented in Ref\cite{dataset7}; consequently, the designation “N.M.” (Not Mentioned) is adopted in the table to denote this absence.

\begin{table*}[!t]
	\centering
	\caption{A Comparative Analysis of Existing Ground-based Cloud Image Datasets. The Proposed CSRC Dataset is a Comprehensive Fine-grained Classification Dataset that Incorporates Both Radiation Source and Color Attributes.}
	{\small
		\begin{tabular}{cccccc}
			\toprule
			Dataset & Fine-grained & Presence of Radiation Source & Color Attributes & Resolution & Sample Size \\
			\midrule
			UTILITY\cite{dataset1}  & No & No & No & 682$\times$512 & 32\\
			SWINSEG\cite{dataset2} & No & No & No & 500$\times$500 & 115\\
			WAHRSIS\cite{dataset3} & No & No & No & 600$\times$600 & 1,013\\
			Fabel et al.\cite{dataset4} & No & No & No & 512$\times$512 & 770\\
			Park et al.\cite{dataset5} & No & No & No & 300$\times$300 & 1,537\\
			TLCDD\cite{dataset6} & No & No & No & 512$\times$512 & 5,000\\
			Kalisch et al.\cite{dataset7} & No & No & No & 3648$\times$2736 & N.M. \\
			Shi et al.\cite{CloudFU-Net} & Yes & No & No & 1704$\times$1704 & 300\\
			CSRC (Ours) & \textbf{Yes} & \textbf{Yes} & \textbf{Yes} & 1260$\times$1260 & 2,330\\
			\bottomrule
	\end{tabular}}
	\label{Table0_1}
\end{table*}

The images in our CSRC dataset were collected at a meteorological observation station located in Xiqing District, Tianjin, China (geographic coordinates: 117.03° E, 39.10° N). The core acquisition device is an All-Sky Imager (ASI-DC-TK02), which provides a wide-angle field of view exceeding 180° and is housed in a waterproof and dustproof enclosure. The system was configured to automatically capture images at fixed 30-second intervals. All images are stored in RGB color JPG format with a resolution of 1260 × 1260 pixels. The CSRC dataset comprises 2,330 samples; the data were randomly shuffled and divided into training, validation, and test sets in a 7:1:2 ratio. To ensure the fidelity of the ground truth, the annotation process was conducted manually by researchers specializing in cloud analysis utilizing the LabelMe software, under the strict supervision of meteorologists. Representative samples from the CSRC dataset are shown in Fig.~\ref{fig_6}.

\subsection{Evaluation Metrics} \label{sec:eva}
As ground-based cloud segmentation fundamentally constitutes a pixel-wise classification task, we employ standard semantic segmentation metrics to quantify model performance. Specifically, we utilize Precision (P), Recall (R), and Mean Intersection over Union (MIoU). All the metrics are derived from true positives (TP), false positives (FP), and false negatives (FN) in experimental results, as specified in the following equations:
\begin{align}
	\mathrm{P} &=\frac{1}{N} \sum \frac{TP_{i}}{TP_{i}+FP_{i}} \\
	\mathrm{R} &=\frac{1}{N} \sum \frac{TP_{i}}{TP_{i}+FN_{i}}\\
	\mathrm{MIoU} &=\frac{1}{N} \sum \frac{p_{i} \cap g_{i}}{p_{i} \cup g_{i}}
	\label{equ_14}
\end{align}\noindent where $C$ represents the number of categories in our dataset, and $i=\left \{ 0, 1, 2, 3 \right \} $ represents a total of four categories.

\subsection{Implementation Details} \label{sec:Imp}
The proposed method was implemented in Python 3.9 using PyTorch 1.10.1 on an Ubuntu 21.04 server equipped with an Intel (R) Xeon (R) Gold 5218N CPU @ 2.30 GHz and an NVIDIA GeForce RTX 4090 GPU (24GB VRAM). Training employed an initial learning rate of $10^{-2}$ with the Adam optimizer, configured with momentum 0.9 and weight decay $10^{-3}$ to ensure stable convergence. All input images were uniformly cropped to 512 $\times$ 512 resolution, consistent with the literature \cite{CloudFU-Net}, to maintain experimental validity. Learning rate decay was applied every 10 epochs, with training termination triggered if no validation loss improvement occurred within 20 consecutive epochs. To enhance model generalization through input diversity, data augmentation strategies including random rotation (0°, 90°, 180°, 270°) and horizontal flipping (probability 0.3) were implemented. The complete training regimen required 8.9 hours for 50 epochs on our dataset.

\subsection{Results and Discussions} \label{Res}
To rigorously evaluate the performance of MPCM-Net, we conducted extensive comparative experiments against a spectrum of baseline methods, ranging from classical semantic segmentation architectures (U-Net \cite{U-Net}, DeepLabv3+ \cite{DeepLabv3}, SegFormer \cite{SegFormer}) to recent, domain-specific SOTA models (CloudFU-Net \cite{CloudFU-Net}, CloudSwinNet \cite{ CloudSwinNet }, MFAFNet \cite{ MFAFNet }, BSANet \cite{ BSANet }, and FA-CloudSeg \cite{ FA-CloudSeg }). To ensure a rigorous and fair comparison, all baseline methods were re-implemented and re-trained specifically on the CSRC dataset. We utilized the official source code provided by the original authors to maintain architectural fidelity. We strictly adhered to a unified experimental protocol, ensuring identical data partitioning, augmentation pipelines, and evaluation metrics across all trials. Furthermore, independent hyperparameter tuning was conducted for each comparative method to guarantee their optimal convergence.

\subsubsection{Quantitative Comparison}
The quantitative evaluation results of the proposed MPCM-Net and other comparative methods on our CSRC dataset are summarized in Table~\ref{Table1}, where the first row lists the adopted pixel-level metrics and subsequent rows present experimental results. The highest scores are boldfaced throughout the table.

\begin{table}[!t]
	\centering
	\caption{Quantitative Comparison with Different Methods on the CSRC Dataset. $\downarrow$ (or $\uparrow$) indicates lower (or higher) is better. The Best Results are Highlighted in Bold.}
	
	{\small
		\begin{tabular}{cccc}
			\toprule
			Method & P($\uparrow $) & R($\uparrow $) & MIoU($\uparrow $) \\
			\midrule
			U-Net (15$'$MICCAI)\cite{U-Net} & 66.7 & 64.6 & 46.7\\
			DeepLabv3+ (17$'$arXiv)\cite{DeepLabv3} & 72.4 & 71.6 & 59.4\\
			SegFormer (21$'$NiPS)\cite{SegFormer} & 84.6 & 85.2 & 61.3\\
			CloudFU-Net (24$'$TGRS)\cite{CloudFU-Net } & 86.9 & 86.5 & 62.1\\
			CloudSwinNet (24$'$Energy)\cite{CloudSwinNet} & 87.1 & 87.6 & 62.3\\
			MFAFNet (25$'$RS)\cite{MFAFNet} & 86.7 & 86.1 & 61.7\\
			BSANet (25$'$RSA)\cite{BSANet} & 86.4 & 86.5 & 61.6\\
			FA-CloudSeg (25$'$ISCAIT)\cite{FA-CloudSeg} & 88.3 & 89.2 & 63.1\\
			MPCM-Net (Ours) & \textbf{90.6} & \textbf{91.1} & \textbf{64.8}\\
			\bottomrule
	\end{tabular}}
	\label{Table1}
\end{table}

Table~\ref{Table1} demonstrates the superior performance of our proposed MPCM-Net over all comparative methods on the CSRC dataset. Specifically, MPCM-Net achieves MIoU improvements of 1.7$\%$, compared to the recent SOTA method. The quantitative results demonstrate that recent methods outperform classical networks through specialized modules. MFAFNet shows a great improvement over the classic methods because the methods employ multi-scale dilated pyramid approaches to enhance the capability to capture multi-scale features. CloudSwinNet introduces a fine-grained feature fusion module into the encoder to capture multi-scale information. However, their reliance on channel concatenation for multi-scale feature fusion lacks effective spatial-channel interaction for global context integration. FA-CloudSeg introduces SA to capture contextual information of different scales in the decoder. CloudFU-Net introduces a content-aware feature recombination module to alleviate the problem of contextual information loss. Their decoders neglect cross-scale spatial-semantic dependencies. In contrast, MPCM-Net's performance advantage stems from its MPC within the MPAC module, which adaptively extracts multi-scale contextual features while enhancing feature interaction, coupled with the MMD that enables comprehensive global-local feature learning through its M2B block.

\subsubsection{Qualitative Comparison}
Qualitative comparisons, presented in Fig.~\ref{fig_7}, provide visual corroboration of the quantitative metrics. We selected challenging samples featuring multi-scale cloud variations, circumsolar occlusion, and complex boundaries. To comparatively evaluate segmentation performance across cloud scales, Fig.~\ref{fig_7} (\uppercase\expandafter{\romannumeral+1}) and (\uppercase\expandafter{\romannumeral+2}) present sequential frames at 30-second intervals from identical regions, while Fig.~\ref{fig_7} (\uppercase\expandafter{\romannumeral+3}) demonstrates performance under diverse meteorological conditions and complex boundaries.
\begin{figure*}[!t]
	\centering
	\includegraphics[width=7.1in]{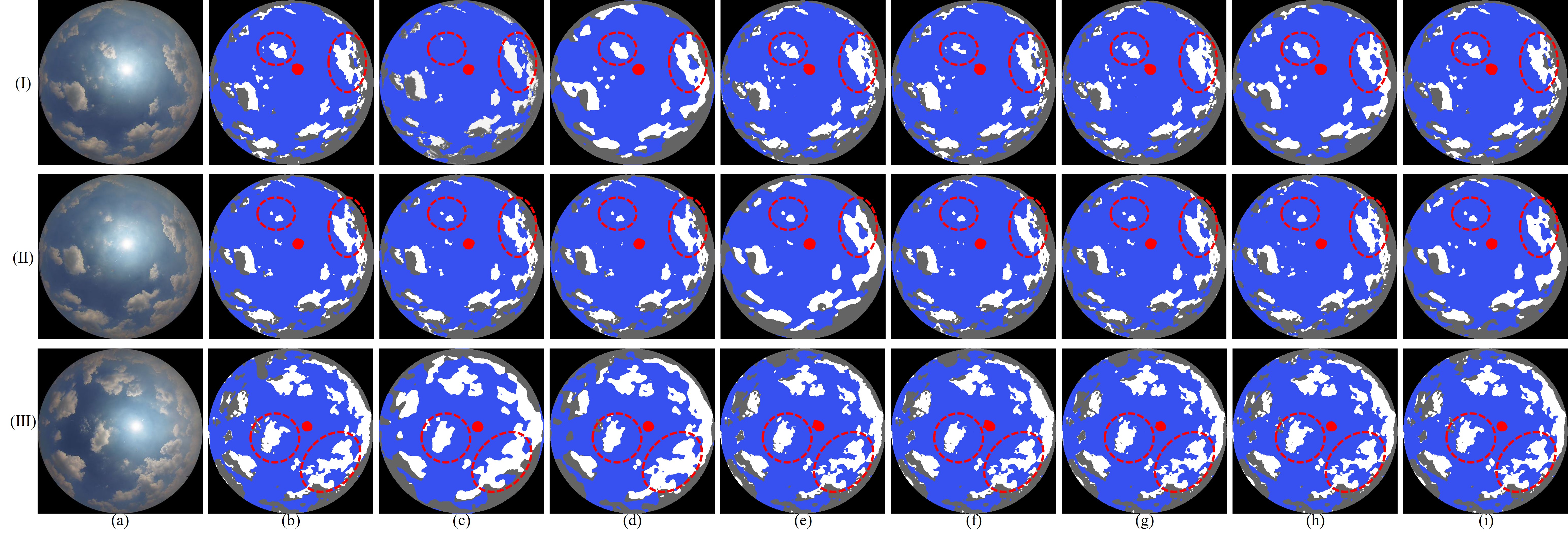}
	\caption{Comparison of segmentation results on the CSRC dataset. (a) Original images. (b) Ground truth. (c) U-Net. (d) Deeplabv3+. (e) SegFormer. (f) CloudFU-Net. (g) MFAFNet. (h) BSANet. (i) Ours. (\uppercase\expandafter{\romannumeral+1}) and (\uppercase\expandafter{\romannumeral+2}) are experimental results of sequential frames from identical regions. (\uppercase\expandafter{\romannumeral+3}) is experimental results of inaccurate boundary segmentation. Among these methods, panels (c)-(e) show classic semantic segmentation algorithms, and panels (f)-(h) show recent ground-based cloud image extraction algorithms. (i) Our method. The key comparison areas are marked with red dashed circles.}
	\label{fig_7}
\end{figure*}

As shown in the first and second rows of Fig.~\ref{fig_7}, cloud formations exhibit significant scale variations across sequential frames. Classical methods (U-Net, DeepLabv3+) frequently omit fine-grained local details of smaller cloud fragments. While CloudFU-Net utilizes adaptive kernel selection to adjust receptive fields, it still exhibits temporal inconsistency when segmenting the same cloud cluster across 30-second intervals. MPCM-Net, leveraging the MPAC module, demonstrates superior consistency, accurately resolving both large-scale stratiform clouds and minute cumulus fragments. 
The third row of Fig.~\ref{fig_7} highlights the challenge of circumsolar regions, where high radiation intensity blurs the spectral boundary between cloud and sky. Traditional algorithms often fail here, misclassifying the aureole as cloud. BSANet and MFAFNet attempt to ameliorate this via dual-branch structures and boundary-attention mechanisms, respectively. However, they occasionally overlook the correlation between local boundary features and the global semantic context. The M2B block with SSHD enhances the network's capability to capture global dependencies among local features, effectively "contextualizing" the ambiguous boundaries and resulting in sharper, more accurate segmentation maps even in saturated circumsolar zones.

\subsubsection{Complexity Comparison}
For practical deployment in ultra-short-term PV power forecasting, inference speed is as critical as accuracy. Table~\ref{Table2} compares the parameters, Floating Point Operations (FLOPs), and inference time of all methods. MPCM-Net establishes an optimal accuracy-efficiency trade-off. It achieves the highest segmentation accuracy while maintaining the lowest inference latency. While our approach does not exhibit a parameter advantage compared to the lightweight BSANet, the significant performance gain fully justifies the additional computational overhead. Notably, it incurs significantly lower computational costs than Transformer-based approaches (FA-CloudSeg, CloudSwinNet). This efficiency is a direct consequence of the M2B architecture's linear scaling law, which avoids the heavy computational burden of the query-key-value calculations inherent in SA mechanisms. Consequently, MPCM-Net is uniquely positioned to meet the real-time requirements of high-frequency ground-based cloud observation systems.

\begin{table}[!t]
	\centering
	\caption{Complexity of Different Comparative Methods on the CSRC Dataset. We Report the Parameters(M), Flops(G), Inference time(ms), and MIoU.}
	
	{\small
		\begin{tabular}{{m{4.14cm}<{\centering}m{0.6cm}<{\centering}m{0.85cm}<{\centering}m{0.6cm}<{\centering}m{0.8cm}<{\centering}}}
			\toprule
			Method & Para($\downarrow$) & FLOPs($\downarrow$) & IT($\downarrow$) & MIoU($\uparrow $) \\
			\midrule
			U-Net (15$'$MICCAI)\cite{U-Net} & 39.5 & 28.7 & 26.9 & 46.7 \\
			DeepLabv3+ (17$'$arXiv)\cite{DeepLabv3} & 40.4 & 30.5 & 27.9 & 59.4 \\
			SegFormer (21$'$NiPS)\cite{SegFormer} & 84.7 & 64.5 & 42.3 & 61.3 \\
			CloudFU-Net (24$'$TGRS)\cite{CloudFU-Net } & 48.8 & 36.7 & 28.7 & 62.1 \\
			CloudSwinNet (24$'$Energy)\cite{CloudSwinNet} & 68.9 & 57.6 & 40.6 & 62.3\\
			MFAFNet (25$'$RS)\cite{MFAFNet} & 34.5 & 23.1 & 16.9 & 61.7\\
			BSANet (25$'$RSA)\cite{BSANet} & 4.29 & 7.5 & 12.5 & 61.6 \\
			FA-CloudSeg (25$'$ISCAIT)\cite{FA-CloudSeg} & 53.6 & 46.4 & 36.8 & 63.1 \\
			MPCM-Net (Ours) & 44.6 & 34.8 & 24.6 & \textbf{64.8} \\
			\bottomrule
	\end{tabular}}
	\label{Table2}
\end{table}

\begin{table}[!t]
	\centering
	\caption{Comparative analysis of partial attention mechanisms ($\mathrm {P}_{*}$) versus full-channel attention mechanisms ($\mathrm {F}_{*}$) on the CSRC dataset.}
	{\small
		\begin{tabular}{ccccc}
			\toprule
			Version & Para($\downarrow$) & FLOPs($\downarrow$) & IT($\downarrow$) & MIoU($\uparrow $)\\
			\midrule
			Baseline & \textbf{44.6} & \textbf{34.8} & \textbf{24.6} & \textbf{64.8}\\
			$\mathrm{F_{C}}$ - $\mathrm{P_{S}}$ - $\mathrm{P_{A}}$ (1) & 46.1 & 35.2 & 25.5 & 64.4\\
			$\mathrm{P_{C}}$ - $\mathrm{F_{S}}$ - $\mathrm{P_{A}}$ (2) & 44.7 & 34.9 & 24.9 & 64.3\\
			$\mathrm{P_{C}}$ - $\mathrm{P_{S}}$ - $\mathrm{F_{A}}$ (3) & 51.3 & 39.7 & 26.9 & 64.7\\
			\bottomrule
	\end{tabular}}
	\label{Table3}
\end{table}

\begin{table}[!t]
	\centering
	\caption{Ablation experiments of ParCM, ParSM, and ParAM on the CSRC dataset.}
	{\small
		\begin{tabular}{ccc}
			\toprule
			Version & Para($\downarrow$) & MIoU($\uparrow $)\\
			\midrule
			Baseline (B) & \textbf{44.6} &\textbf{64.8}\\
			B w/o ParCM (1)  & 44.3 & 63.6\\
			B w/o ParCM+ParSM (2)  & 43.1 & 62.5\\
			B w/o ParCM+ParSM+ParAM (3) & 39.2 & 61.6\\
			\bottomrule
	\end{tabular}}
	\label{Table4}
\end{table}

\subsection{Ablation Study}
To validate the efficacy of key components in our proposed MPCM-Net, comprehensive ablation studies were conducted. Built upon an encoder-decoder architecture, MPCM-Net incorporates the MPAC in the encoder and the Multi Mamba decoding. The MPAC consists of MPC and MPA blocks, primarily comprising multi-scale partial attention modules (ParM, ParSM, and ParAM). The M2B within the decoder is fundamentally constructed using SSHD. Therefore, we conduct multiple ablation configurations on the CSRC dataset to verify module effectiveness. Note that MPCM-Net served as the baseline method in all experiments.
\begin{table}[!t]
	\centering
	\caption{Comparison experiments of CA and SLA with SE and SA on the CSRC dataset. Baseline is our method, Row1 uses SE instead of CA, Row2 uses SA instead of SLA.}
	{\small
		\begin{tabular}{cccc}
			\toprule
			Version & Para($\downarrow$) & IT($\downarrow$) & MIoU($\uparrow $)\\
			\midrule
			Baseline & \textbf{44.6} & \textbf{24.6} & \textbf{64.8}\\
			SE (1) & 45.4 & 24.9 & 64.1\\
			SA (2) & 47.2 & 26.1 & 64.3\\
			\bottomrule
	\end{tabular}}
	\label{Table5}
\end{table}
\begin{figure}[!t]
	\centering{\includegraphics[width=3.5in]{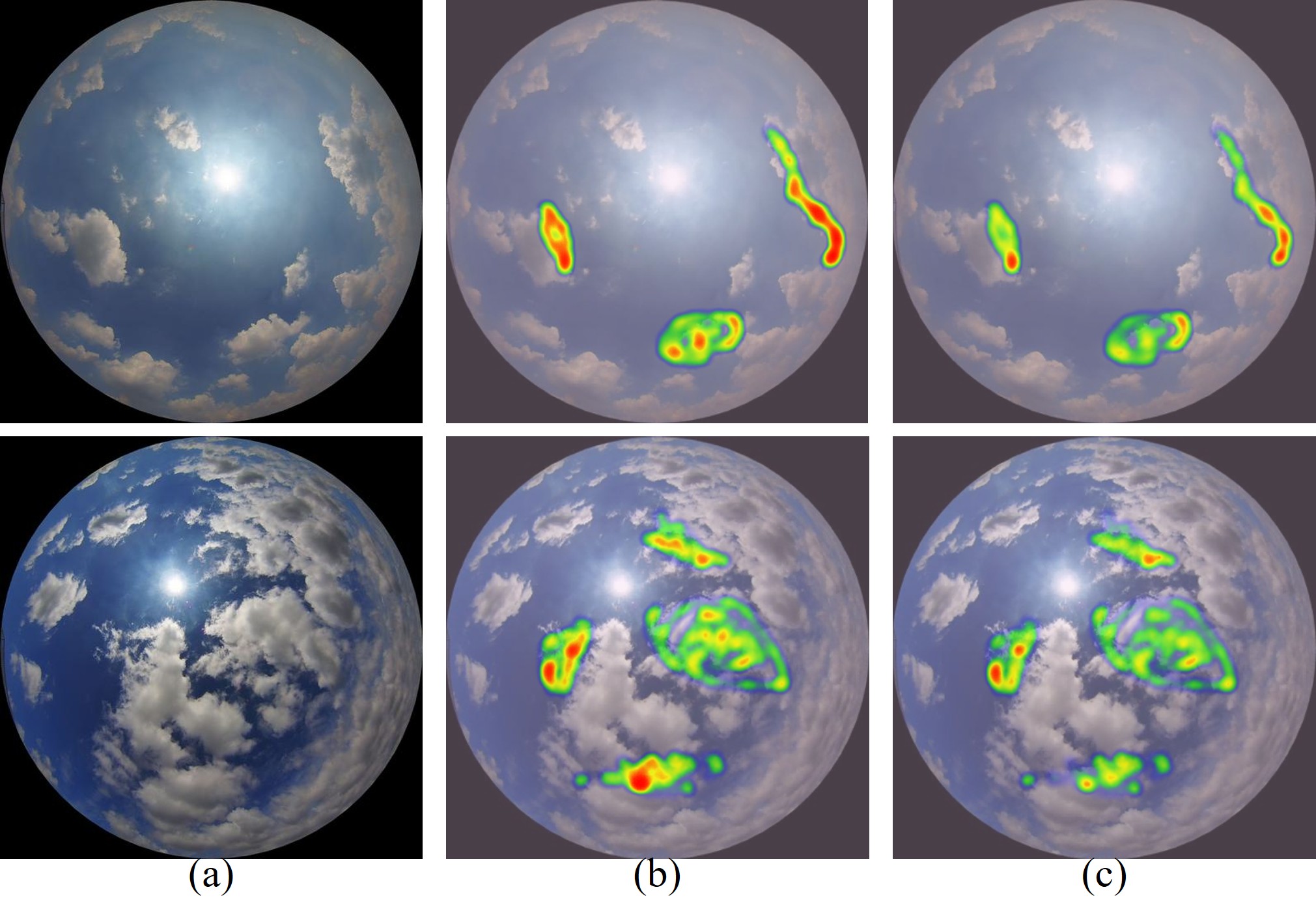}}
	\caption{Visualization results show different categories of the CSRC dataset validation set using Grad-CAM. (a) Input image, (b) MPCM-Net with partial attention mechanism, and (c) MPCM-Net with full-channel counterparts.}
	\label{fig_8}
\end{figure}

We first investigated the advantage of the proposed partial attention mechanism over conventional full-channel attention. By replacing the partial modules ($\mathrm{P_{C}}$, $\mathrm{P_{S}}$, and $\mathrm{P_{A}}$) with their full-channel counterparts ($\mathrm{F_{C}}$, $\mathrm{F_{S}}$, and $\mathrm{F_{A}}$), we benchmarked the model against traditional visual attention paradigms as presented in Table~\ref{Table3}.
The results indicate that partial attention configurations consistently outperform full-channel baselines in both accuracy and speed. Specifically, the partial variants achieved MIoU improvements of 0.4$\%$, 0.5$\%$, and 0.1$\%$, alongside inference latency reductions of 0.9 ms, 0.3 ms, and 2.3 ms, respectively. This suggests that processing a subset of representative channels or spatial regions reduces information redundancy while maintaining feature discriminability. GradCAM visualizations (Fig.~\ref{fig_8}) corroborate this, showing that the partial attention mechanism directs the network's focus more precisely onto cloud boundaries and fine-grained textures.

We further dissected the MPAC by incrementally removing the ParCM, ParSM, and ParAM modules from the baseline according to the configurations in Table~\ref{Table4}. Results demonstrate that all three proposed modules consistently enhance model performance. Specifically, comparing the baseline with Row 1 reveals 1.2$\%$ accuracy degradation without ParCM. This indicates that the partial channel module strengthens spatial information interaction, improving segmentation accuracy. Comparisons between the baseline and Rows 2-3 show 1.1$\%$ and 0.9$\%$ accuracy reductions without ParSM and ParAM, respectively. These results confirm that the partial spatial module effectively acts as a spatial descriptor, adaptively focusing on relevant regions, while the partial attention module captures essential long-range dependencies.

\begin{figure}[!t]
	\centering{\includegraphics[width=3.5in]{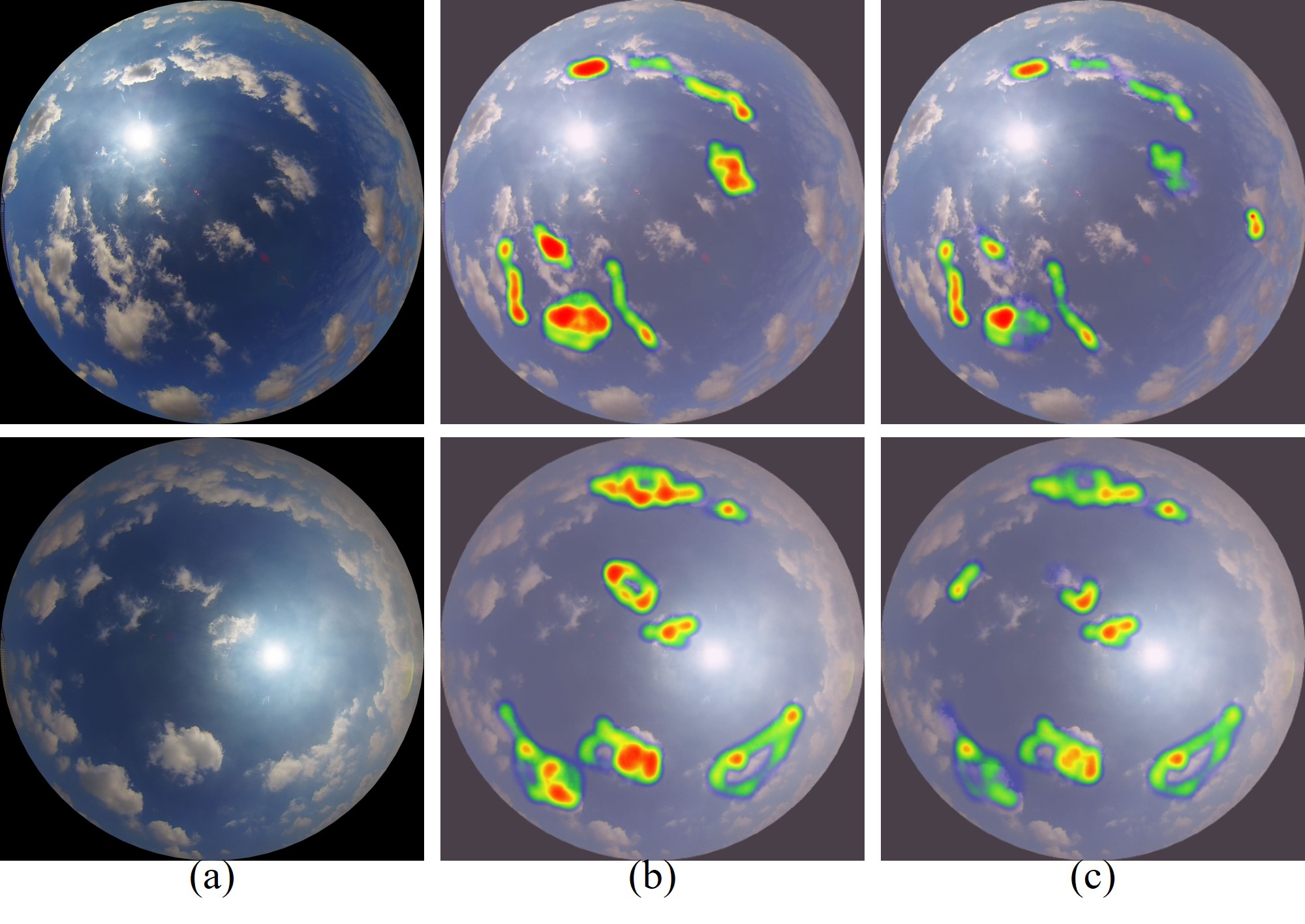}}
	\caption{Visualization results show different categories of the CSRC dataset validation set using Grad-CAM. (a) Input image, (b) MPCM-Net, and (c) MPCM-Net with SE.}
	\label{fig_9}
\end{figure}

Within the ParCM and ParAM modules, we integrated CA and SLA to replace standard SE-Net and SA mechanisms. Table~\ref{Table5} and Fig.~\ref{fig_9} demonstrate that CA and SLA provide superior performance. The CA-based model achieved a 0.7$\%$ MIoU gain over the SE-Net baseline, attributed to CA's ability to encode precise positional information, which is a crucial factor for segmenting clouds with random motion and multi-scale variations. Likewise, SLA demonstrated superior efficiency, leveraging the benefits of linear complexity to process spatial information faster than standard SA mechanisms.
\begin{table}[!t]
	\centering
	\caption{Ablation experiments of MPAC and MMD on the CSRC dataset.}
	{\small
		\begin{tabular}{cccc}
			\toprule
			Version & MPAC & MMD & MIoU($\uparrow $)\\
			\midrule
			Baseline &  \checkmark & \checkmark & \textbf{64.8}\\
			1 &   & \checkmark & 61.2\\
			2 & \checkmark &  & 61.6\\
			\bottomrule
	\end{tabular}}
	\label{Table6}
\end{table}
\begin{figure}[!t]
	\centering{\includegraphics[width=3.5in]{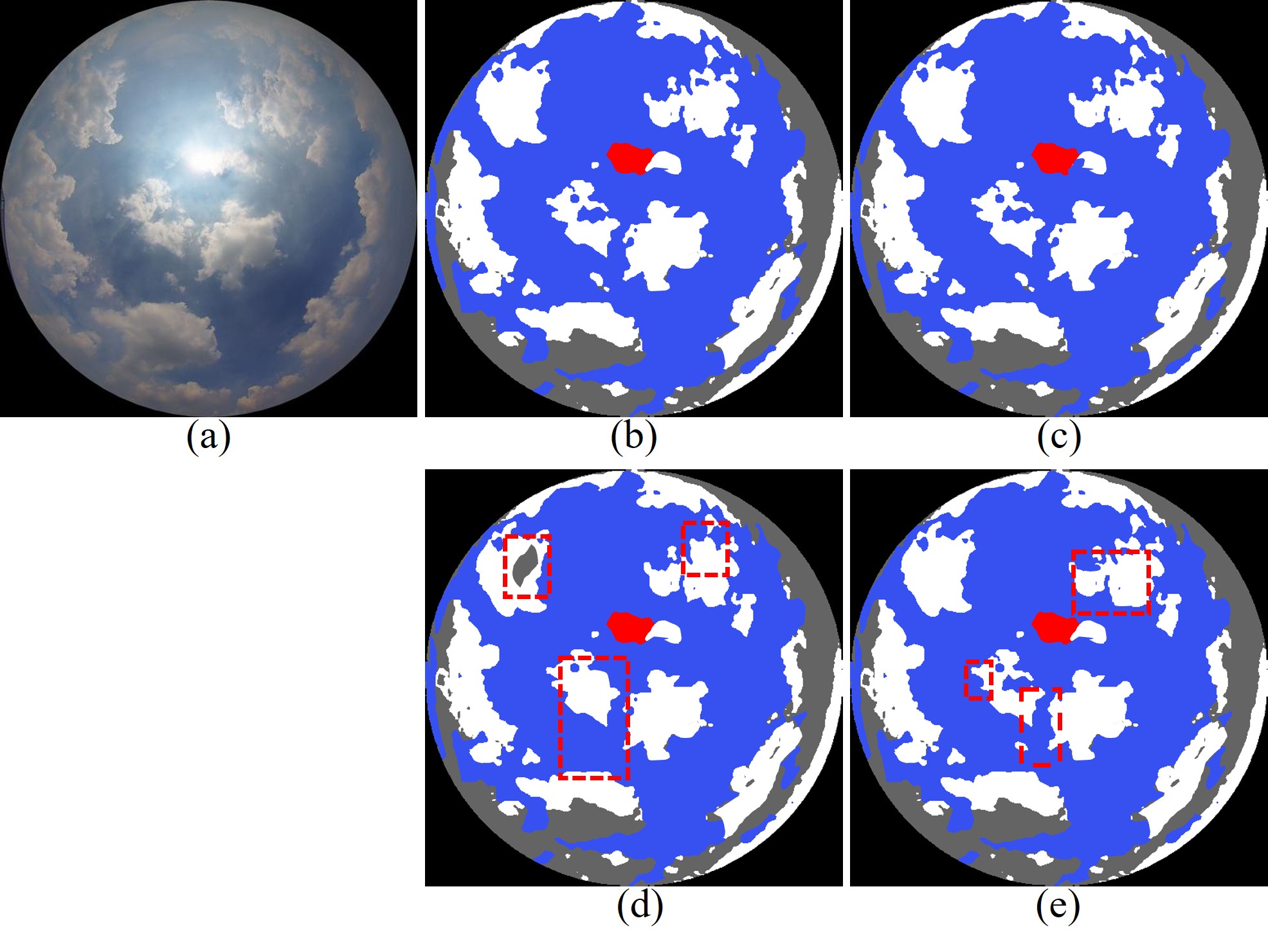}}
	\caption{The segmentation results of the ablation experiments. The results are shown in (a) original images, (b) ground truth, (c) MPCM-Net, (d) MPCM-Net without MPAC, and (e) MPCM-Net without MMD, respectively. The red rectangles in column (d) indicate the misclassification of scale variable cloud clusters in the results of removing MPAC. The red rectangles in column (e) indicate the misclassification of the boundary information in the results of removing MMD. Column (c) shows that our method can effectively alleviate these problems.}
	\label{fig_10}
\end{figure}

Finally, we assessed the holistic impact of the MPAC and MMD (Table~\ref{Table6} and Figure~\ref{fig_10}). It should be noted that ResNet-50 is employed as the encoder in the configuration without MPAC. The removal of the MPAC module resulted in a substantial 3.6$\%$ drop in MIoU, leading to poor segmentation of multi-scale cloud clusters (The comparison between (c) and (d) of Fig.~\ref{fig_10}). This validates MPAC's capability in multi-scale context extraction. Conversely, excluding the MMD (and its M2B block) caused a 3.2$\%$ performance decline, characterized by significant degradation in boundary integrity as shown in Fig.~\ref{fig_10} (e). This underscores the MMD's vital role in establishing long-range correlations between the encoder and decoder, effectively recovering spatial details lost during down-sampling and ensuring precise boundary delineation.

\section{Conclusion}\label{sec:conclusion}
The precise ground-based cloud image segmentation constitutes a fundamental prerequisite for grid dispatch operations and clean energy integration. To surmount the inherent limitations in existing DL approaches by proposing MPCM-Net, this article introduces a novel framework integrating partial convolution in the encoder and Mamba architecture in the decoder. Our principal contributions are threefold: First, we designed a novel MPAC encoder that facilitates interaction between local and global features across diverse cloud scales through ParCM and ParSM. Second, we introduced the SLA mechanism, a high-performance attention variant that yields comprehensive multi-scale feature representation. Third, we developed a M2B in the decoder that effectively mitigates contextual information loss through spatial-semantic multi-dimensional feature integration. Extensive experimentation on our newly proposed CSRC dataset demonstrates the significant superiority of MPCM-Net over SOTA methods. Comprehensive ablation studies rigorously validate the effectiveness of each module. The results confirm that our MPCM-Net achieves a superior equilibrium between segmentation accuracy and inference speed for ground-based cloud segmentation. Future work will focus on extending MPCM-Net to ultra-short-term PV power forecasting tasks, further enhancing its practical utility in renewable energy management systems.

 
%

\bibliographystyle{IEEEtran}
\bibliography{reference}

\newpage

\begin{IEEEbiography}[{\includegraphics[width=1in,height=1.25in,clip,keepaspectratio]{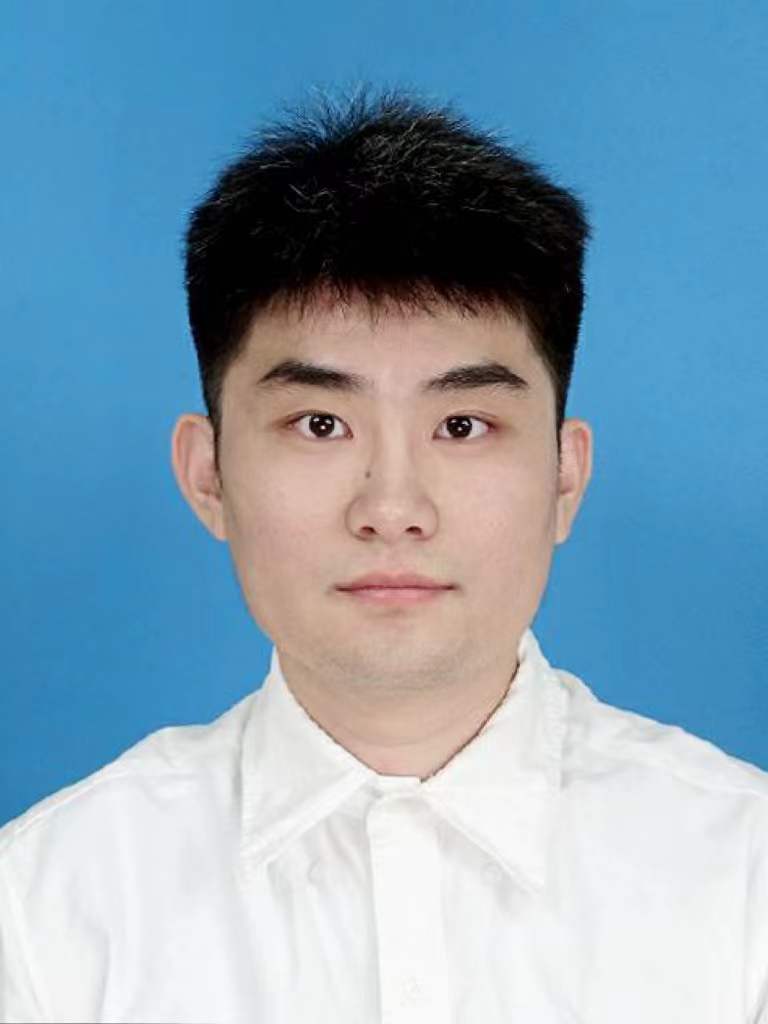}}]{Penghui Niu}
	received the B.S. degree in automobile engineering from Shanghai Dianji University, Shanghai, China, in 2017 and the M.S. degree in instruments science and technology from Tianjin University of Science and Technology, Tianjin, China, in 2021. He is currently pursuing the Ph.D. degree in control science and engineering with the School of Artificial Intelligence, Hebei University of Technology, Tianjin, China. His research interests include ultra-short-term PV power forecasting and intelligent interpretation of remote sensing images.\end{IEEEbiography}

\begin{IEEEbiography}[{\includegraphics[width=1in,height=1.25in,clip,keepaspectratio]{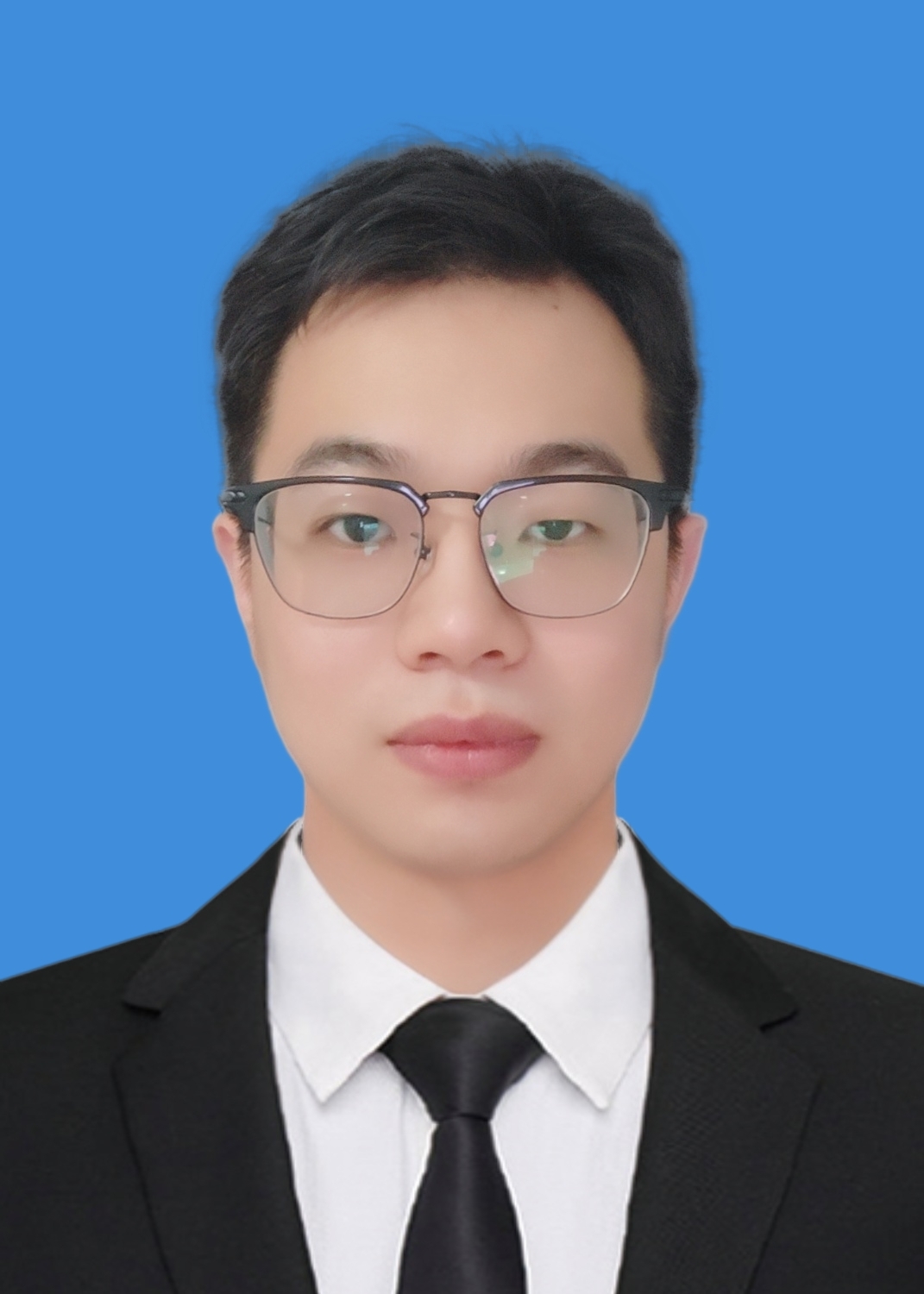}}]{Jiashuai She}
	received the B.S. degree in computer science and technology from Hebei University of Technology, Tianjin, China, in 2022. He is currently pursuing the Ph.D. degree in electrical engineering and automation with the School of Electrical Engineering, Hebei University of Technology, Tianjin, China. His research interests include ultra-short-term PV power forecasting and intelligent interpretation of remote sensing images.\end{IEEEbiography}

\begin{IEEEbiography}[{\includegraphics[width=1in,height=1.25in,clip,keepaspectratio]{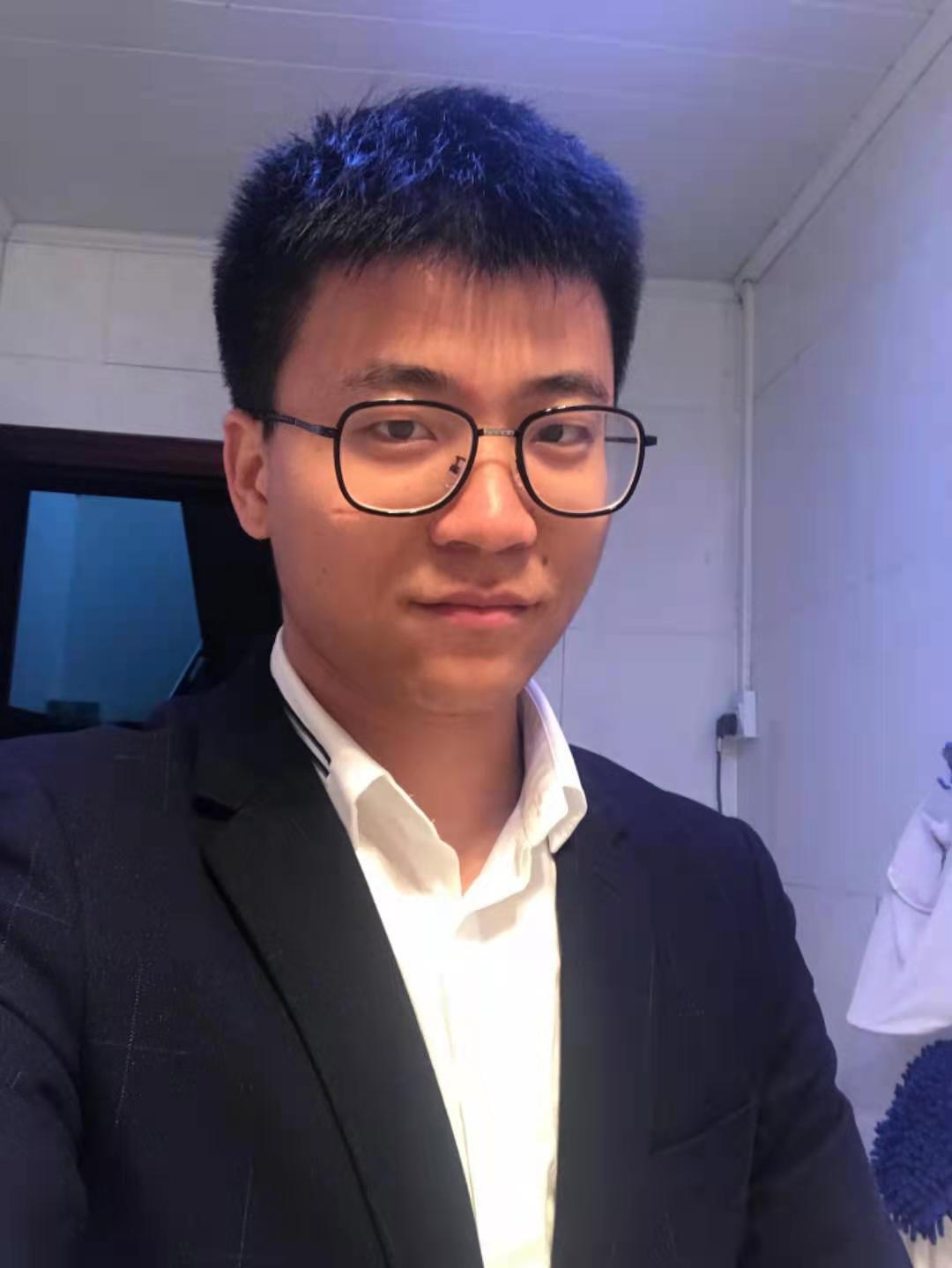}}]{Taotao Cai}
	received the PhD degree from Deakin University, in 2020 after spending two years with the University ofWestern Australia (2017-2019) and over one year at Deakin University (2019-2020) during his Ph.D. studies. 
	
	He is a lecturer in computing with the University of Southern Queensland (UniSQ). His primary focus is on research and teaching in the field of Data Science, including graph data processing, social network analytics, recommendation systems, and complexity science. Prior to joining the faculty at UniSQ, Taotao held positions as a PostDoctoral Research Fellow at Macquarie University (2021-2023) and an associate research fellow at Deakin University (2020-2021). During this time, he made significant research contributions, which have been published in leading international conferences and journals such as IEEE ICDE, Information Systems, and IEEE TKDE.\end{IEEEbiography}

\begin{IEEEbiography}[{\includegraphics[width=1in,height=1.25in,clip,keepaspectratio]{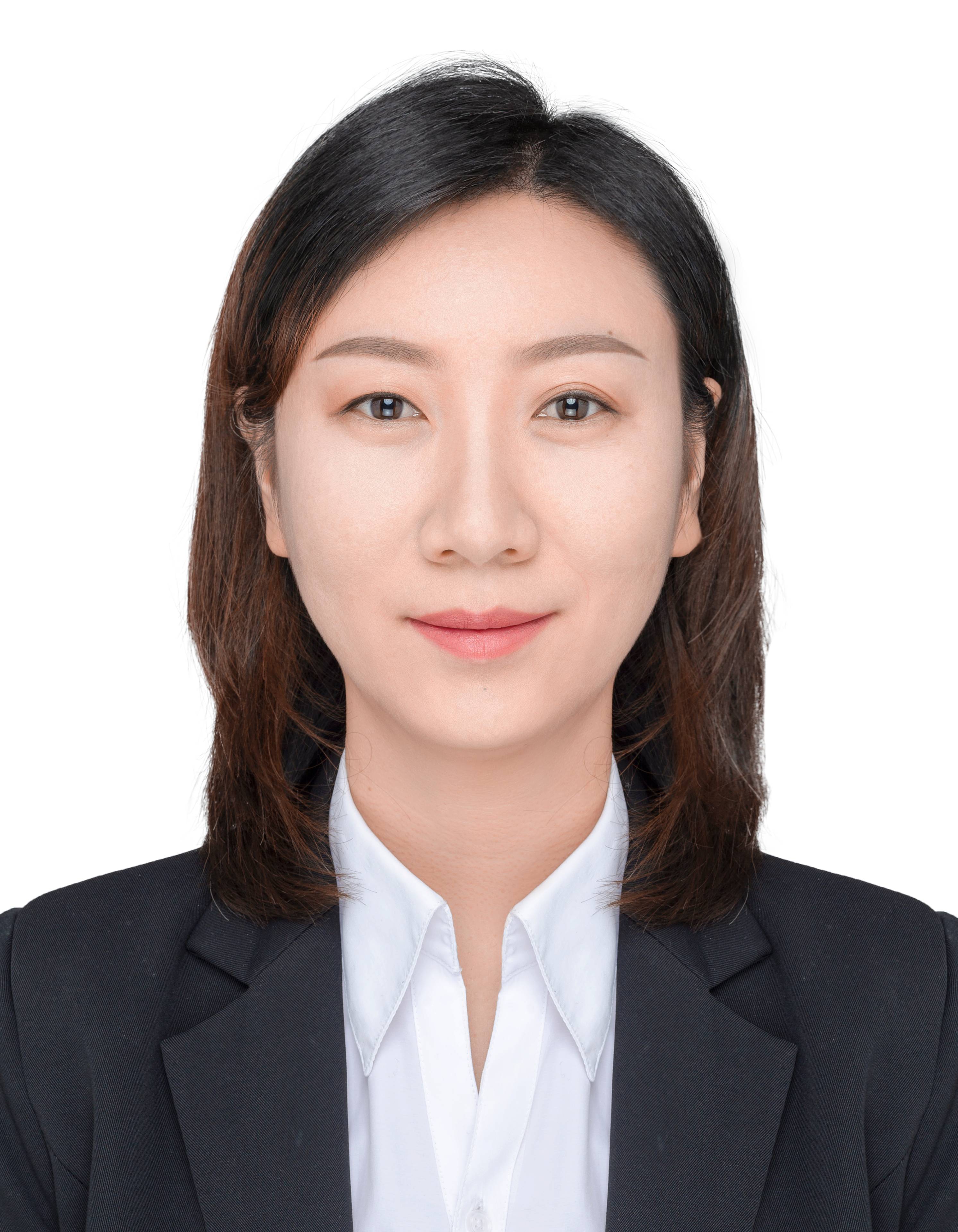}}]{Yajuan Zhang}
	received the B.S. degree in computer science and technology from Inner Mongolia Agricultural University, Inner Mongolia, China, in 2007 and the  M.S. degree in control science and engineering from Hebei University of Technology, Tianjin, China, in 2010. 
	
	She is currently a Senior Experimentalist in the School of Artificial Intelligence at Hebei University of Technology. Her current research interests include image processing and data mining.\end{IEEEbiography}

\begin{IEEEbiography}[{\includegraphics[width=1in,height=1.25in,clip,keepaspectratio]{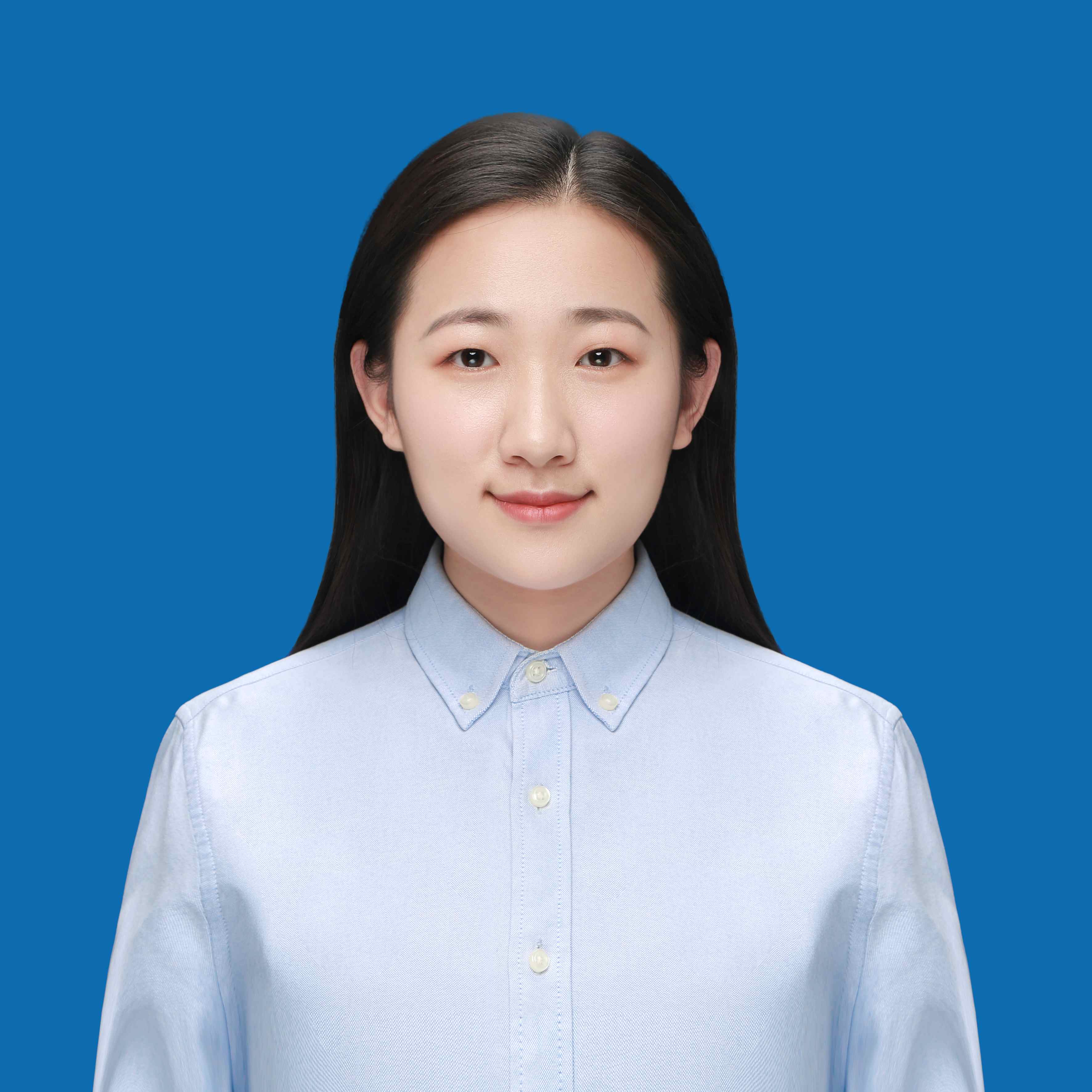}}]{Ping Zhang}
	received the M.S. degree in software engineering from the College of Software, Jilin University, Changchun, China, in 2018, and the Ph.D. degree in computer science from Jilin University, Changchun, in 2021.
	
	She is currently a lecturer at the School of Artificial Intelligence, Hebei University of Technology, Tianjin, China. Her research interests include feature selection and machine learning.\end{IEEEbiography}	

\begin{IEEEbiography}[{\includegraphics[width=1in,height=1.25in,clip,keepaspectratio]{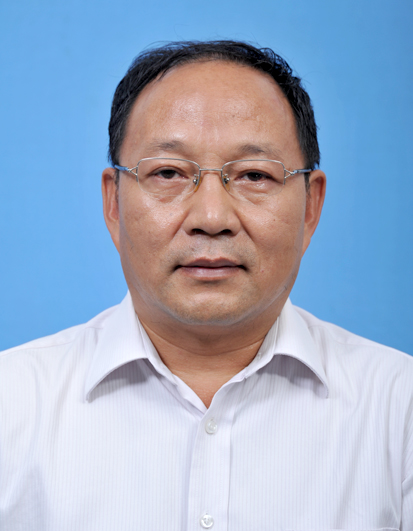}}]{Junhua Gu}
	received the B.S. degree in mathematics from Shanghai Jiaotong University, Shanghai, China, in 1988, and the M.S. degree in computer science and the Ph.D. degree in electrical engineering from the Hebei University of Technology, Tianjin, China, in 1993 and 1997, respectively.
	
	He is a Professor with the School of Artificial Intelligence, Hebei University of Technology. He is the Hebei new century “333 Talent Project” second level suitable person. He has published more than 70 papers and the current research interests include big data, intelligent control, and intelligent transportation systems.\end{IEEEbiography}
	
\begin{IEEEbiography}[{\includegraphics[width=1in,height=1.25in,clip,keepaspectratio]{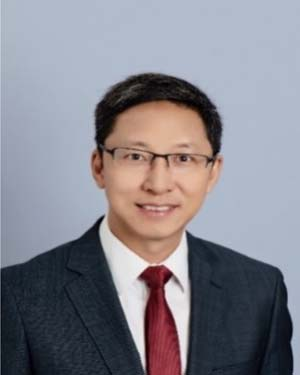}}]{Jianxin Li}
	(Senior Member, IEEE) received the PhD degree in computer science from the Swinburne University of Technology, Melbourne, VIC, Australia, in 2009. He is a professor of Information Systems with the School of Business and Law, Edith Cowan University, Joondalup, Australia.
	
	He was awarded as World Top 2$\%$ scientists by 2023 Stanford due to his research impact and high citation. His current research interests include database query processing and optimization, social network analytics, and traffic network data processing. He is a senior member of the IEEE Computer Society. He has published about 200 peer-reviewed and high-quality research articles in top international conferences and journals, including SIGMOD, PVLDB, ICDE, ACM WWW, SIGKDD, ACM CIKM, IEEE TKDE, KBS, TII, IS, etc. There are 12 research articles that were awarded as the Highly Cited Papers or Hot Papers in 2020-2024. Currently, he serves as the Editor-in-Chief in Array, Associate Editors in Information Systems, Knowledge-based Systems, and IEEE Signal Processing Letters, World Wide Web Journal. As Program Committee Chairs or Steering Committee Members, he chaired the international conferences of ADMA’2019, APWeb-WAIM’2023, and more than 10 workshops since 2014. \end{IEEEbiography}

\vfill

\end{document}